\documentclass[final,5p,times,twocolumn]{elsarticle}

\usepackage{amssymb}
\usepackage{amsthm}

\usepackage[switch]{lineno}

\usepackage{hyperref}
\usepackage{xspace}
\usepackage{multirow}
\usepackage{amsmath}
\usepackage[dvipsnames]{xcolor}
\usepackage{algorithm}
\usepackage{flushend}
\usepackage{adjustbox}
\usepackage{float}
\usepackage{letltxmacro}
\usepackage{xparse}
\usepackage{algpseudocode}
\usepackage[list=true]{subcaption}
\newcommand{\cross}{~\!~\!$\times$\!~\!~}
\newcommand{\around}{$\sim$~\!}

\hypersetup{
    colorlinks,
    citecolor=black,
    filecolor=black,
    linkcolor=black,
    urlcolor=black
}

\NewDocumentCommand{\subautoref}{om}{%
    \IfValueTF{#1}{%
        \hyperref[#2]{\autoref*{#2}#1}%
    }{%
        \autoref{#2}%
    }%
}

\journal{arXiv}

\begin{document}

\begin{frontmatter}



\title{Unified Dynamic Scanpath Predictors Outperform Individually Trained Neural Models}


\author[UHH]{Fares Abawi}
\author[UHH,UniS]{Di Fu}
\author[UHH]{Stefan Wermter}

\affiliation[UHH]{
            department={Department of Informatics,},
            organization={University of Hamburg},
            addressline={Vogt-Koelln-Str. 30},
            postcode={22527}, 
            state={Hamburg},
            country={Germany}}

\affiliation[UniS]{
            department={School of Psychology,},
            organization={University of Surrey},
            addressline={Elizabeth Fry Building (AD)},
            city={Guildford},
            state={Surrey},
            country={United Kingdom}}
          
\begin{abstract}
Previous research on scanpath prediction has mainly focused on group models, disregarding the fact that the scanpaths and attentional behaviors of individuals are diverse. The disregard of these differences is especially detrimental to social human-robot interaction, whereby robots commonly emulate human gaze based on heuristics or predefined patterns. However, human gaze patterns are heterogeneous and varying behaviors can significantly affect the outcomes of such human-robot interactions. To fill this gap, we developed a deep learning-based social cue integration model for saliency prediction to instead predict scanpaths in videos. Our model learned scanpaths by recursively integrating fixation history and social cues through a gating mechanism and sequential attention. We evaluated our approach on gaze datasets of dynamic social scenes, observed under the free-viewing condition. The introduction of fixation history into our models makes it possible to train a single unified model rather than the resource-intensive approach of training individual models for each set of scanpaths. We observed that the late neural integration approach surpasses early fusion when training models on a large dataset, in comparison to a smaller dataset with a similar distribution. Results also indicate that a single unified model, trained on all the observers' scanpaths, performs on par or better than individually trained models. We hypothesize that this outcome is a result of the group saliency representations instilling universal attention in the model, while the supervisory signal and fixation history guide it to learn personalized attentional behaviors, providing the unified model a benefit over individual models due to its implicit representation of universal attention.
\end{abstract}




\end{frontmatter}


\section{Introduction}
\begin{figure*}[!t]
\centering
\includegraphics[width=0.98\linewidth]{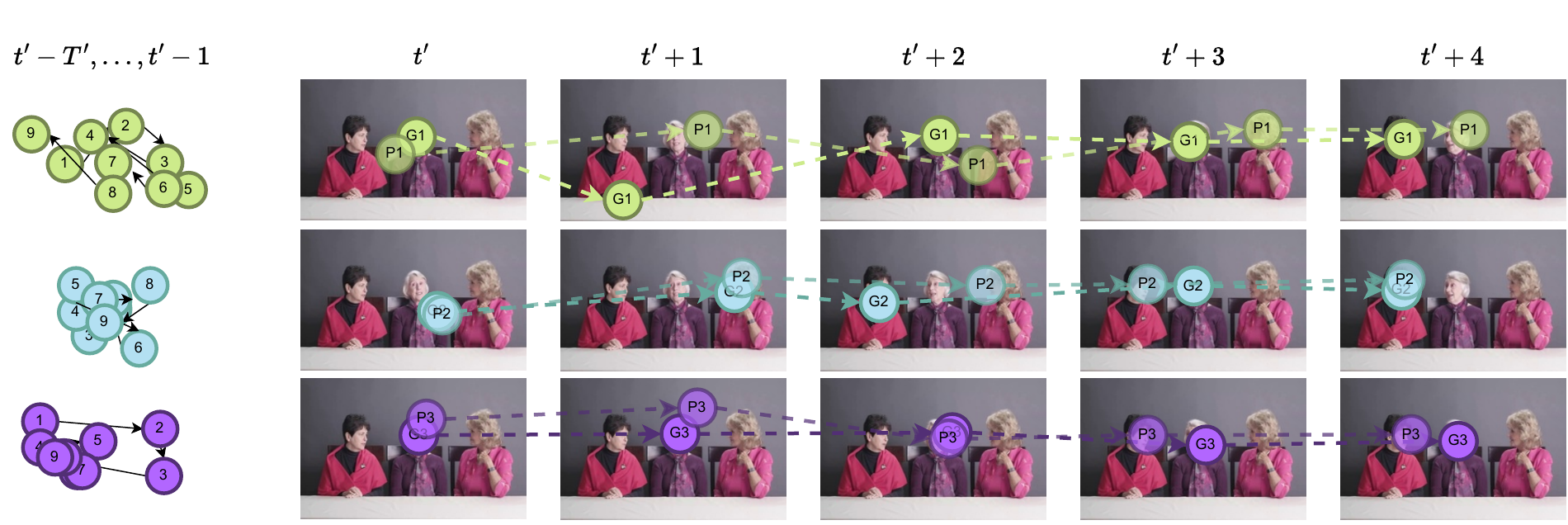}
\caption{\textcolor{black}{Our modification of the GASP~\citep{abawi2021gasp} model, transforming it into a scanpath prediction model by including a fixation history module. The GASP model integrates the facial expression and gaze direction social cue representations with the DAVE~\citep{tavakoli2020deep} model's fixation density maps. By additionally integrating the fixation history as another modality, we are able to specify the observer whose scanpath is to be inferred. We provide a sequence with a specific context size (number of input frames) consisting of prior fixations and spatiotemporal representations. In this example, we employ the late integration GASP variant (DAM + LARGMU, context size $T'=10$) predicting scanpaths of three observers (G1 - G3). At each timestep of multi-step-ahead (five steps) fixation, predictions (P1 - P3) indicate that the slightest divergence from the ground-truth has a noticeable impact on future predictions.}}
\label{fig:multi-step-ahead}
\end{figure*}
Gaze shapes and guides social interactions, both for signaling and perceiving intent~\citep{canigueral2019role,salley2016conceptualizing}. Similarities across human eye movement patterns are described as \textit{universal attention}~\citep{xu2017beyond} and are attributed to memory effects, bottom-up saliency, oculomotor biases, and physical constraints~\citep{li2017individual}. However, gaze patterns are influenced by socio-ecological factors and behavioral traits, that could differ depending on the observer. These factors contrive \textit{personalized attention}~\citep{xu2017beyond}.

In this study, we focus mainly on modeling human gaze patterns, known as \textit{scanpaths}, to better simulate the cognitive behaviors exhibited during social interactions. The ability to simulate scanpaths is especially necessary for conducting human-robot interaction studies, where the gaze of the robot could greatly impact humans' perception and social acceptability of it~\citep{belkaid2021mutual,lathuiliere2019neural}. \textcolor{black}{Such a model would also enable robots to mimic multiple human scanpath patterns for conducting cognitive robot simulations, without requiring human participants~\citep{fu2023congruence}. Moreover, such a model could be relevant to directing users' attention during video conferencing, adapting to their scanpaths.} 

We model scanpaths under the free-viewing condition, whereby the observer is instructed to freely watch a video, without any predetermined objectives or tasks. The uniqueness of the free-viewing condition lies in the fact that it does not require any explicit gaze target. The viewing patterns under this condition are comparable to those exhibited by animals when foraging for food~\citep{bartumeus2009optimal,damelio2021gazing}. This implies a universal goal that can be associated with all observers. However, deviations in gaze patterns from the norm can be attributed to the intrinsic motivation of observers, shaped by their personalized attention~\citep{caligiore2015intrinsic}. 

Our task addresses the learning of scanpaths on observing dynamic (video) visual scenes. A closely related task is saliency prediction~\citep{tavakoli2020deep,xiong2023casp}, in which attention maps are learned based on the gaze of multiple observers. With minor modifications, saliency models can be used as predictors of scanpaths~\citep{kuemmerer2021scanpath,fu2023congruence}. Saliency prediction models represent statistical measures of fixation distributions, visualized as spatial attention maps.  However, equating the peaks of these maps with the fixation targets throughout a sequence might not accurately reflect an individual's scanpath. This could lead to abrupt transitions between fixations.

One potential solution is to train or fine-tune saliency models to predict individual scanpaths, maintaining traits emerging as a result of universal and personalized attention. However, this approach could be hindered by the sparsity of the predicted fixation maps (hereafter denoted as \textit{priority maps}), particularly if there's no input signal maintaining a record of fixation sequences for a single observer. Nevertheless, saliency predictors have shown great effectiveness in modeling human attention and could provide features pertaining to universal attention, therefore, useful for predicting scanpaths. While both saliency and scanpath prediction are closely related, there are critical differences that necessitate modifying the former task to better address the latter. We motivate the need for modifications based on the following observations:

\bigskip

\noindent \textbf{Saliency models predict group attention, whereas scanpath models predict the sequential attention of an individual}. Namely, saliency models predict the distribution of fixation probabilities for multiple observers, whereas scanpath models predict the fixation trajectory of an individual observer viewing a scene. To repurpose a saliency model for predicting scanpaths, we need to fine-tune the model for each observer independently in order to represent their unique viewing patterns. To circumvent this limitation, we could alternatively train a single unified model by additionally providing a prior that can separate different scanpath trajectories. By focusing on individual scanpaths, we can represent the prototypical personalized attention patterns.

\noindent \textbf{Scanpaths of observers are non-deterministic and unique to the individual}, meaning that they can vary from one viewing of a scene to the next. Previous research has shown that repetition of trials can increase the similarity between subsequent scanpaths for each observer, suggesting that humans have a natural tendency to follow different scanpaths without prior exposure to a given stimulus~\citep{foulsham2008can}. This phenomenon highlights the importance of using sequential models that can maintain previous context, such as a memory component for storing the fixation history. This is in contrast to saliency prediction models, which infer fixation distributions that remain unchanged given the same stimuli. 

\noindent \textbf{Salient stimuli and low-level features influence scanpaths}. Unlike saliency prediction, scanpath prediction models the unique fixations of a single observer. Jiang~et~al.~\cite{jiang2016learning} have identified several factors that can influence the uniqueness of individual scanpaths, including low-level visual features, semantics, central bias, and fixation shift distribution. Based on these findings, we propose a sequential audiovisual scanpath prediction model that can implicitly represent personalized patterns. More concretely, our model is not designed to explicitly represent these factors, but it can infer them from the input features, such as social cue representations and attention maps. This allows our model to infer the patterns of attention that are characteristic of human observers.

Training a separate \textit{individual} model for each observer could lead to optimal prediction of their scanpaths. However, this approach is prohibitively expensive given a large number of observers. We investigate whether individual models are necessary as opposed to a single \textit{unified} model, distinctively predicting each scanpath based on the individual fixation histories. We pose the following questions:
    \begin{enumerate}
        \item Are the fixation histories adequate priors for differentiating scanpath trajectories?
        \item Does a model trained on individual viewing patterns independently, yield better predictions of scanpaths compared to a unified model trained on all fixations?
        \item How many multi-step-ahead fixations in a sequence can be reliably predicted from a given scanpath and stimuli before model predictions diverge (illustrated in~\autoref{fig:multi-step-ahead})?
    \end{enumerate}

\noindent To tackle these questions, we propose a scanpath prediction model and framework that allows for the exploration of each. Given that social cues play an important role in modulating visual attention~\citep{hessels2020does}, we utilize GASP~\citep{abawi2021gasp}, a dynamic saliency prediction model employing sequential gating mechanisms to augment raw audiovisual samples with representations of social cues. \textcolor{black}{During social interaction, observers look at human faces 79\% of the time~\citep{xu2018findwho}, indicating that features relating to faces are of great importance, especially in conversational settings. Two of the most visual attention-directing features in faces relate to facial expressions (affect-biased attention)~\citep{todd2012affect,pritsch2017perception} and the gaze direction of the observer (gaze cueing of attention)~\citep{frischen2007gaze}. We, therefore, employ those features in our current model, similar to the GASP model~\citep{abawi2021gasp}.} The best-performing GASP model concatenates the social cue representations, weighing their contributions by emphasizing weaker ones, and finally, combining them using an attentive convolutional LSTM~\citep{cornia2018predicting} followed by a gating module~\citep{arevalo2020gated}. We discard the gaze following social cue representation since it is shown to have an insignificant or even detrimental effect on the performance of GASP. Moreover, we extend the model with a fixation history channel that maintains a fixed number of previous fixation masks---a sequence of fixation points blurred by a 2D Gaussian filter with width equating to a $1^{\circ}$ viewing angle.  

In summary, we introduce a framework for modeling and evaluating dynamic scanpaths, inspired by existing methodologies in scanpath prediction for images~\citep{kuemmerer2021scanpath}. Additionally, we present a modular multimodal architecture, designed for flexibility in accommodating various modules for detecting social cues. A key aspect of our approach is the utilization of an observer's fixation history, enabling the model to learn scanpaths of multiple observers using a single unified model.

\section{Related Work}
\label{related_work}
In this work, we adapt a saliency prediction model~\citep{abawi2021gasp} for individual observers' scanpath prediction. Since interacting with and perceiving stimuli within social settings requires the encoding of sequential knowledge, we provide an overview of the existing literature on both saliency and scanpath prediction, with an emphasis on dynamic variants, i.e., models designed for inferring either task on videos. Static variants are designed to operate on images and therefore,  are out of the scope of this work.

\subsection{Audiovisual Dynamic Saliency Prediction}
Auditory and multimodal features have become a target of interest in the context of visual saliency modeling~\citep{tavakoli2020deep, boccignone2018give, min2020multimodal, tsiami2020stavis}. Tavakoli~et~al.~\cite{tavakoli2020deep} propose a simple deep learning model based on 3D-ResNet~\citep{hara20183dresnet} for encoding the visual (video) and auditory streams separately. The two streams are joined, reduced in dimensionality, and decoded as two-dimensional priority maps with high intensity in regions where participants (observers) tend to fixate. Tsiami~et~al.~\cite{tsiami2020stavis} propose early fusion of audiovisual representations. At multiple levels of the visual stream, they introduce supervised attention modules. At a deep stage of the visual stream, the visual features guide the auditory stream to the most salient regions, resulting in a more accurate localization of sound sources. Eventually, the supervised attention modules and the auditory features are concatenated, after which their representations are reduced to a single two-dimensional feature map. Min~et~al.~\cite{min2020multimodal} propose a multistage audiovisual saliency model that minimizes the discrepancy between the proposed locations arriving from auditory and visual modalities. The modalities generate spatiotemporal saliency maps, which are adaptively fused in the final stage. Both early fusion~\citep{tsiami2020stavis,min2020multimodal} approaches for integrating auditory and visual features are shown to outperform late fusion~\citep{tavakoli2020deep} methods.

Jain~et~al.~\cite{jain2020avinet} present a 3D convolutional encoder-decoder model for predicting saliency. They explore different audiovisual fusion techniques and show that introducing auditory input does not result in significant improvement to the performance of their model. Dynamic saliency models relying only upon visual stimuli tend to perform on par with audiovisual models~\citep{wang2021spatio,bellitto2021hierarchical,droste2020unifiedia} supporting the finding of Jain~et~al.~\cite{jain2020avinet}. The consensus on whether auditory stimuli play a significant role in predicting saliency is not clear. Yang~et~al.~\cite{yang2023svgc} introduce an audiovisual graph convolution-based model, emphasizing the importance of multimodal input for predicting saliency in 360$^{\circ}$ videos. Xiong~et~al.~\citep{xiong2023casp} argue that audiovisual saliency models may underperform due to temporal inconsistencies between auditory and visual streams. Despite some studies indicating visual-only models might suffice, our study primarily focuses on social attention, which inherently depends on both visual and auditory stimuli. Consequently, we rely on an audiovisual saliency model~\citep{tavakoli2020deep} in our framework.

\subsection{Dynamic Scanpath Prediction}
Scanpath prediction on dynamic scenes requires accommodating changes to stimuli along with the modeling of fixation trajectories. Most of the existing research addresses the prediction of egocentric gaze in videos recorded using head-mounted cameras. Li~et~al.~\cite{li2013learning} present a model for learning the temporal dynamics in first-person activity videos, utilizing motion and pose features relating the head and hand of the actor. Huang~et~al.~\cite{huang2018predicting} propose a multitask model for predicting saliency and task-guided attention transitions using independent 3D-CNN streams. Zhang~et~al.~\cite{zhang2017deep} predict gaze on the future frames which are generated using a GAN. An identical discriminator model receives the future observations and anticipated future frames produced by the 3D-CNN generator. Concurrently, an independent 3D-CNN predicts fixations on the generated frames. 

Most of these models are goal-directed, whereby the objective is known and the gaze fixations are supervisory signals. However, Aakur~and~Bagavathi~\citep{aakur2021unsupervised} address egocentric gaze prediction as an unsupervised task. Their model is separated into three stages, initially extracting appearance and motion features, followed by a symbolically represented stage indicating the direction of information flow between spatial regions in the video. Finally, the model generates an attention map indicating the predicted fixation corresponding to locations with maximum energy. 

Another line of research addresses the prediction of the egocentric scanpath under the free-viewing condition~\citep{xu2018gaze,xu2019predicting,naas2020functional,xu2022spherical,randon2022track,li2023scanpath}. Xu~et~al.~\cite{xu2018gaze} train their model on individual observer fixation trajectories while freely viewing 360$^{\circ}$ videos on a VR headset. The model receives a video frame at a given timestep, concatenated with local and global spatiotemporal saliency features. A recurrent model encodes the fixation trajectories and its latent representation is concatenated with the visual encoding of all saliency features to predict the displacement in fixation for the next frames. Naas~et~al.~\cite{naas2020functional} follow a relatively similar approach, replacing local features with an optical flow representation. These approaches predict the scanpaths of a single observer for each video given their past fixation histories as priors. However, all aforementioned architectures are trained to model the scanpaths within a limited viewport~\citep{li2023scanpath}, unique to each observer. Consequently, the learned universal attention patterns are applicable to any individual, rather than representing their personalized attention.

Scanpath prediction on social videos is a less explored domain. Coutrot~et~al.~\cite{coutrot2018scanpath} propose a generalizable framework for predicting and classifying scanpaths based on a Hidden Markov Model (HMM) and discriminant analysis. Their approach is examined on static natural scenes and dynamic social scenes, identifying three location states for the HMM. These states are then used for classifying information relating to the observers or the stimuli. Boccignone~et~al.~\cite{boccignone2020gaze} rely on multimodal social cues as priors to a stochastic model, simulating the fixation patch transitions as a Poisson process. Lan~et~al.~\cite{lan2022eyesyn} design a psychologically-inspired model for synthesizing gaze. Their method addresses the detection of actions, including ``verbal communication'', based on simulated gaze. Such approaches represent universal behaviors but do not address individual differences among multiple observers. 

To alleviate this gap, we design models that understand and predict the personalized gaze patterns of individuals. By doing so, we can evaluate the magnitude of these differences and determine whether the uniqueness in gaze patterns necessitates the development of personalized models, tailored for each individual observer.

\section{Methods}
\begin{figure*}[!hbtp]
\centering
\subcaptionbox{Early Fusion (DAM + ARGMU)}{\includegraphics[width=0.47\textwidth, trim = 2em 0 26.8em 0, clip=true]{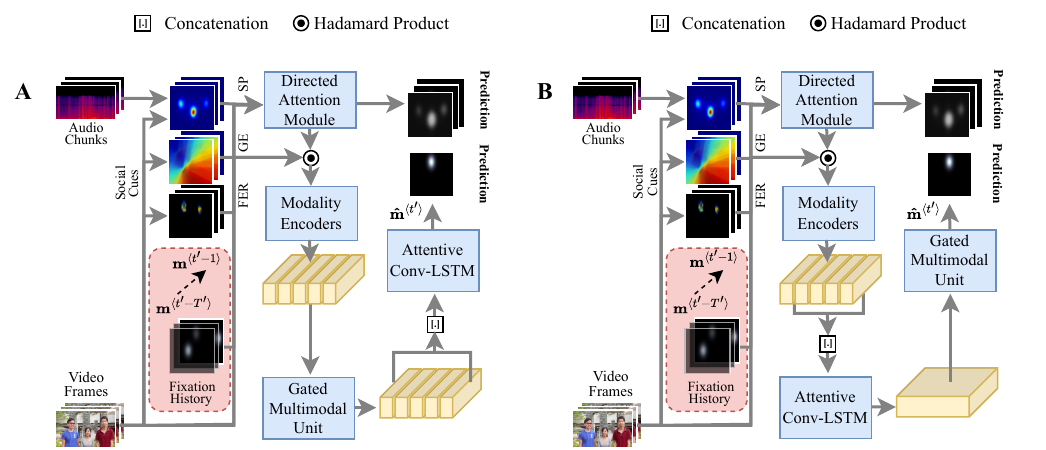}}%
\hfill
\subcaptionbox{Late Integration (DAM + LARGMU)}{\includegraphics[width=0.47\textwidth, trim= 26.8em 0 2em 0, clip=true]{src/img/model_variants4.pdf}}%
\caption{\textcolor{black}{Our two GASP~\citep{abawi2021gasp} variants extended with fixation history modules for predicting scanpaths, where \textbf{(a)} is the modality \textit{fusion} variant ARGMU, and \textbf{(b)} is the non-fusion late \textit{integration} model LARGMU. The directed attention module (DAM) is applied to each variant with the fixation density maps for the entire sequence as ground-truth during training. $T'$ represents the context size (number of input frames) for each model, whereas $t'$ indicates the current timestep (frame index) in the video. $\mathbf{\hat m}^{\langle t' \rangle}$ represents the priority map predicted by the model at timestep $t'$. \\ \footnotesize{ SP:~Saliency Prediction Representation;~GE: Gaze Direction Estimation Represention; FER:~Facial Expression Recognition Representation.}}}
\label{fig:model_variants}
\end{figure*}
\textcolor{black}{In previous work~\citep{abawi2021gasp}, we introduced GASP, a social attention model that augments saliency prediction models with social cue representations to improve performance in social settings. The model is composed of two stages, starting with the extraction and representing of social cues, followed by feature integration. In the first first stage, spatiotemporal representations are extracted using pretrained social cue detectors. Namely, the model infers gaze direction and facial expressions from a sequence of video frames, represents those features as images, and integrates them with the fixation density maps of an audiovisual saliency prediction model. In the second stage, the social cue representations and the fixation density maps are integrated, following one of two paradigms depending on the model variant. The late integration variant combines the social cue and saliency representations into a single branch after integrating them sequentially. Conversely, the early fusion variant combines the representations prior to sequential integration. The early fusion variant requires fewer parameters yet often under-performs its late integration counterpart with the same context size (number of input frames). An overview of the two variant structures is illustrated in~\autoref{fig:model_variants}.}

\textcolor{black}{Separating the feature representation and saliency modeling into two stages is both biologically plausible and computationally efficient. From a psychological standpoint, our approach follows the feature integration theory~\citep{treisman1980feature}, which states that low- and high-level features are processed in an initial stage. During this stage, only features of all objects are extracted, since prior knowledge about the relevance of an object is not yet processed. In a second stage, the features are clustered into objects, and each object is assigned a relevance, allowing for selectivity in attention toward the most conspicuous one. From a computational perspective, each social cue and saliency representation can be computed in parallel given that features relating to the interactions between different modalities are not required. The lower bound in terms of time complexity, is determined by the slowest detection and representation modality.}

\textcolor{black}{In this work, we modify the GASP~\citep{abawi2021gasp} model for predicting scanpaths and adopt the two-stage approach proposed by Abawi~et~al.~\cite{abawi2021gasp}. This modification entails the addition of a fixation history module to enable distinguishing between observers. Additionally, the directed attention module and the sequential integration model should consider the fixation history module as an additional modality. Moreover, the most critical modification to the GASP model for converting it to a scanpath predictor relates to the means by which we train the model. In contrast to saliency prediction where samples are trained on the fixation density map of a group of observers, our scanpath predictor is trained such that each sample has a ground-truth priority map indicating the gaze target for a single observer.}

\textcolor{black}{In this section, we provide a detailed overview of the GASP components, along with the additional fixation history module, making the transformation from a saliency to a scanpath prediction model possible.}

\subsection{Sampling and Social Cue Detection}
We represent social cues following the paradigm introduced by Abawi~et~al.~\cite{abawi2021gasp}. We retain DAVE~\citep{tavakoli2020deep} as the saliency predictor, Gaze360~\citep{kellnhofer2019gaze360} as the gaze estimator, and the facial expression recognizer developed by Siqueira~et~al.~\cite{siqueira2020efficient}. We discard the gaze following modality~\citep{recasens2017following} due to its high time complexity and insignificant improvement to GASP~\citep{abawi2021gasp}.

During the fine-tuning phase, the image captures are downsampled to 10 frames per second. This aligns with the finding that eye fixations change within an interval of 100 to 500~ms~\citep{rayner1998eye}. The frames are pushed to a queue with a maximum size matching that of the modality with the longest context: DAVE with a context size of 16 visual frames.

Auditory signals are resampled to 16~kHz for accommodating videos irrespective of their original sampling rate. Resampling requires audio recordings of at least one second to avoid introducing artifacts. During training, we split one-second recordings beginning with the first visual frame in the context window into 16 chunks. We then extract 64 bands of the log mel-spectrogram with overlapping windows of .025s having a hop length of .01s following the same preprocessing technique adopted by Tavakoli~et~al.~\cite{tavakoli2020deep}. The resulting coefficients are propagated to the auditory stream of DAVE.

\subsection{Fixation History Module}
\label{sec:fix_hist}
For predicting scanpaths of individual observers, the model requires a mechanism for recalling previous fixations. This becomes relevant considering the scanpath differs for each observer exposed to the same stimuli and their scanpaths are dependent on their previous fixation points. During training, the fixation history is set to a sequence of 2D priority maps, created by applying Gaussian blur on the fixation points. The Gaussian blur filter's width corresponds to a $1^{\circ}$ viewing angle as a function of distance from the display monitor. This filter is applied to the previous ground-truth fixation points preceding the last timestep for a given sample. This translates to the teacher forcing strategy~\citep{williams1989learning} during training i.e., the ground-truth maps of previous timesteps are fed as model inputs to predict the map of the current timestep. 

During evaluation, our model predicts a priority map indicating the target of attention for an individual observer. The maps are then queued in the fixation history. The previous fixations define the context of attention, enforcing a foveated region upon the different input modalities and assisting in the prediction of the next fixation for an observer. The overall process equates to scanpath prediction with the added benefit of operating on dynamic contexts given an arbitrary number of timesteps. For predicting scanpaths, the fixation history cannot be discarded, especially in a unified model, as it serves as the primary mechanism for distinguishing between scanpath trajectories.

In~\autoref{alg:exec_pipeline}, we present the scanpath evaluation pipeline for each observer. The context size $T'$ defines the number of recurrent timestep representations arriving from the different cue detectors. The predicted fixation $\mathbf{\hat{m}}^{\langle t'\rangle}$ at timestep $t'$ is fed back into the fixation history for an arbitrary number of multiple steps ahead. We note the model's performance is primarily evaluated based on the output from its first prediction step. This output reflects the model's initial predictions, without extending into multi-step-ahead evaluations. To evaluate a model's capability in handling extended sequences without relying on the ground-truth after initializing the fixation history, we can iteratively input the model's predictions into the fixation history queue. This approach resembles the detection pipeline, with the key distinction of not acquiring the fixation history from ground-truth for subsequent stimuli detection. Evaluating multiple steps ahead allows us to assess the model's accuracy in forecasting future steps, closely reflecting real-world scenarios where we don't have access to the ground-truth data.
\begin{algorithm}[!ht]
\small
\caption{The dynamic scanpath evaluation pipeline incorporating the fixation history, similar to K\"ummerer~and~Bethge~\cite{kuemmerer2021scanpath}}\label{alg:exec_pipeline}
\begin{algorithmic}[1]
\State \textbf{Definitions:}
\State \hspace{0.1cm} vf: video frames, \hspace{0.1cm} ac: audio chunks
\State \hspace{0.1cm} fh: fixation history
\State \hspace{0.1cm} DetectCues: detect social cues
\State \hspace{0.1cm} Shift: shift left and discard first element
\State \hspace{0.1cm} UpdateFrame: update last video frame
\State \hspace{0.05cm} UpdateChunks: update audio chunks if new 1s sample reached
\State \hspace{0.1cm} Integrate: sequential integration
\State \hspace{0.1cm} Sample: get video or audio at specified rate
\State \hspace{0.1cm} Eval: evaluate saliency metrics
\State
\State $t' \gets \text{current sub-sampled video frame index}$
\State $\text{vf} \gets \text{Sample(16 frames, 10 fps)}$
\State $\text{ac} \gets \text{Sample(1s audio, 16 chunks)}$
\State $t' \gets t' + 16$
\For {$t'' \in \text{\{1,~\dots~, context size $T'$\}}$}
    \State $\text{cues}[t''] \gets \text{DetectCues(vf, ac)}$
    \State $\text{fh}[t''] \gets \mathbf{m}^{\langle t'\rangle}$
    \State $\text{vf} \gets \text{Shift(vf) \& UpdateFrame(vf)}$
    \State $\text{ac} \gets \text{UpdateChunks(ac)}$
    \State $t' \gets t' + 1$
\EndFor
\For {$n \in \text{\{0,~\dots~, multi-step-ahead predictions\}}$}
    \State $\mathbf{\hat{m}}^{\langle t'\rangle} \gets \text{Integrate(cues, fh, vf[16 - $T'$:16])}$
    \State $\text{Eval}(\mathbf{\hat{m}}^{\langle t'\rangle}, \mathbf{m}^{\langle t'\rangle})$
    \State $\text{vf} \gets \text{Shift(vf) \& UpdateFrame(vf)}$
    \State $\text{ac} \gets \text{UpdateChunks(ac)}$
    \State $\text{cues} \gets \text{Shift(cues)}$
    \State $\text{cues}[T'] \gets \text{DetectCues(vf, ac)}$
    \State $\text{fh} \gets \text{Shift(fh)}$
    \State $\text{fh}[T'] \gets \mathbf{\hat{m}}^{\langle t'\rangle}$
    \State $t' \gets t' + 1$
\EndFor
\end{algorithmic}
\end{algorithm}

\subsection{Sequential Integration Model}
We describe the components of the sequential integration method presented in GASP~\citep{abawi2021gasp}. The attentive recurrent gating mechanism as well as its late integration variant are detailed in this work. We also provide an overview of the directed attention module's role in improving a model's performance.

\subsubsection{Directed Attention Module}
The Directed Attention Module (DAM)~\cite{abawi2021gasp} is based on the Squeeze-and-Excitation~\citep{hu2018squeeze} model for extracting the channel-wise interactions between the input modalities. The number of channels $C'$ is defined by $K$\cross$C$ where $K$ is the total number of modalities and $C$ denotes the number of image channels per feature map, assuming all modalities have an equal number of image channels and dimensions. The initial aggregation in the form of average pooling across the channel pixels is expressed as follows:
\begin{equation}
    \label{eq:squeeze}
    \boldsymbol{\ell}^{[1]} = \frac{1}{H\!~\!\times\!~\!W} \sum_{h=1}^{H} \sum_{w=1}^{W} \mathbf{r}_{c'}(h,w),
\end{equation}
where $\boldsymbol{\ell}^{[1]}$ represents the squeeze operation, $H$ and $W$ represent the height and width of the feature maps respectively, whereas $\mathbf{r}_{c'}$ signifies the standardized feature map channel representation. The aggregated representations of all channels are then compressed and expanded using two linear fully-connected layers, with a non-linear activation following the first:
\begin{equation}
    \label{eq:excitation}
    \boldsymbol{\ell}^{[2]} = \sigma (\mathbf{W}_{\ell}^{[2]} \cdot \text{relu}(\mathbf{W}_{\ell}^{[1]} \cdot \boldsymbol{\ell}^{[1]})),
\end{equation}
where $\boldsymbol{\ell}^{[2]}$ is the non-linear channel weight vector for scaling the channel contribution. The Sigmoid activation function $\sigma$ is used as a gating mechanism rather than an attention mechanism, simply to avoid having a single active channel as would be the case if it were Softmax (attention) instead. The compression and expansion layer parameters are $\mathbf{W}_{\ell}^{[1]}$ and $\mathbf{W}_{\ell}^{[2]}$ respectively, where  \mbox{$\mathbf{W}_{\ell}^{[1]} \in \mathbb{R}^{\frac{C'}{\gamma}\!~\!\times\!~\!C'}$} and \mbox{$\mathbf{W}_{\ell}^{[2]} \in \mathbb{R}^{C'\!~\!\times\!~\!\frac{C'}{\gamma}}$}. The reduction ratio $\gamma$ is a hyperparameter controlling the factor by which the channels are compressed. Finally, each feature map $\mathbf{r}_{c'}$ is scaled by its corresponding gain arriving from the $\boldsymbol{\ell}^{[2]}$ layer. 

Following the approach presented by Abawi~et~al.~\citep{abawi2021gasp}, we duplicate the Squeeze-and-Excitation model, denoting one by the direct stream and the other by the inverted stream. The former has a direct path to the scanpath model, whereas the inverted stream has a separate output head. The output head is composed of a 2D convolutional layer with 32 channels, a kernel shape of 3\cross3, and a padding of 1, followed by a max-pooling layer with a window size of 2\cross2, effectively reducing the feature map by half its size. The final layer aggregates the pooled representation to a single channel by applying 1\cross1 convolution.

The direct stream receives the concatenated channels of the social cue, saliency prediction, and fixation history representations. The final weighted feature maps (target maps) of the direct stream are propagated to our scanpath model. The number of target maps corresponds to the number of channels received by the direct stream. As the name implies, the inverted stream acquires the chromatically inverted modality representations by applying a non-linear transformation \mbox{$\mathbf{r}^{-1}_{c'}\!=\!-\text{softmax}(\mathbf{r}_{c'})$} to the modality channels. The output of the inverted stream predicts a sequence of fixation density maps (attention maps), corresponding to universal attention learning. These fixation density maps represent the top-down and bottom-up attention of multiple observers, which is prior knowledge that the individual does not possess. However, the plausibility of a bottom-up saliency detector is evaluated based on its resemblance to fixations of multiple individuals, when the task is designed to minimize top-down effects~\citep{seo2009static}. 

We assume that these attention maps represent the ideal saliency maps since a clear separation cannot be formed between bottom-up and top-down attention. Moreover, we hypothesize that individual differences in attention should be large enough to distinguish between the scanpath trajectories. This implies that extrinsic factors that attract attention would result in the highest fixation density per frame on average.

The motivation behind introducing the directed attention module lies in avoiding bias toward the saliency prediction representation, being both an input and target of the model. As a result, autoencoding the saliency input would be the optimal outcome, reducing all other modality connection parameters to zero. A more performant saliency model would amplify the biased reliance on its representation, leading to better performance overall, however, the generalization suffers. This is evident from the observation that training models on biased datasets leads to incorrect feature learning, albeit successful on the provided samples, a phenomenon in deep neural networks known as shortcut-learning~\citep{geirhos2020shortcut}. To address this bias, the model emphasizes weaker modality representations, i.e., modality representations that have a low spatial match to the ground-truth maps. Stronger representations are inhibited, causing the model to assign a larger gain weight to their representations during the learning phase. However, this is the case only when weaker representations lack information content that is sufficient to guide loss minimization.

\subsubsection{Modality Encoders}
The modality encoders are 2D convolutional neural networks with a structure similar to encoder-decoder models, i.e., feature compression of visual modalities through a bottleneck followed by decompression. We follow the same encoder structure described in GASP~\cite{abawi2021gasp} to initialize or model parameters with those of the pretrained models. 

\subsubsection{Attentive Convolutional LSTM and the Gated Multimodal Unit}
The Attentive Convolutional LSTM (ALSTM)~\citep{cornia2018predicting} is an adaption to the convolutional LSTM model for recursively attending to feature maps. This structure has demonstrable advantages over conventional recurrent convolutional models and proves effective in modeling saliency as illustrated by Cornia~et~al.~\cite{cornia2018predicting}. A convolutional LSTM is expressed as follows:
\begin{equation}
    \label{eq:alstm}
    \begin{aligned}
    \mathbf{i}^{\langle t\rangle} &= \sigma(\mathbf{W}_i * \mathbf{s}^{\langle t\rangle} + \mathbf{U}_i * \mathbf{h}^{\langle t-1\rangle} + b_i ), \\
    \mathbf{f}^{\langle t\rangle} &= \sigma(\mathbf{W}_f * \mathbf{s}^{\langle t\rangle} + \mathbf{U}_f * \mathbf{h}^{\langle t-1\rangle} + b_f ), \\
    \mathbf{o}^{\langle t\rangle} &= \sigma(\mathbf{W}_o * \mathbf{s}^{\langle t\rangle} + \mathbf{U}_o * \mathbf{h}^{\langle t-1\rangle} + b_o ), \\
    \mathbf{g}^{\langle t\rangle} &= \tanh{(\mathbf{W}_c * \mathbf{s}^{\langle t\rangle} + \mathbf{U}_c * \mathbf{h}^{\langle t-1\rangle} + b_c)}, \\
    \mathbf{c}^{\langle t\rangle} &= \mathbf{f}^{\langle t\rangle} \odot \mathbf{c}^{\langle t-1\rangle} + \mathbf{i}^{\langle t\rangle} \odot \mathbf{g}^{\langle t\rangle}, \\
    \mathbf{h}^{\langle t\rangle} &= \mathbf{o}^{\langle t\rangle} \odot \tanh({\mathbf{c}^{\langle t\rangle}}), \\
    \mathbf{q}^{\langle t\rangle} &= \mathbf{W}_q * \tanh{(\mathbf{W}_a * \mathbf{s}^{\langle t\rangle} + \mathbf{U}_a * \mathbf{h}^{\langle t-1\rangle} + b_a)},
    \end{aligned}
\end{equation}
where $\mathbf{W}_i$, $\mathbf{W}_f$, $\mathbf{W}_o$, and $\mathbf{W}_c$, represent the kernel parameters of the input $\mathbf{i}^{\langle t\rangle}$, forget $\mathbf{f}^{\langle t\rangle}$, output $\mathbf{o}^{\langle t\rangle}$ gates, and cell state respectively. The bias unit for each projection layer is denoted by $b_i, b_f,  b_c,$ and $b_o$. The input map $\mathbf{s}^{\langle t\rangle}$ is convolved with all gate parameters at each timestep. The cell state is denoted by $\mathbf{c}^{\langle t\rangle}$ and the hidden state by $\mathbf{h}^{\langle t\rangle}$. The hidden state is convolved with the recurrent parameters $\mathbf{U}_i$, $\mathbf{U}_f$, $\mathbf{U}_o$, depending on the projection layer to which they apply. The convolution kernels are of size 3\cross3, with a padding of 1, having 32 channels each. We note that $\mathbf{q}^{\langle t\rangle}$ represents the pre-attentive output of the model at each timestep. The pre-attentive output has separate $\mathbf{W}_q$ kernel parameters that are convolved with the activated input map and previous hidden state. The input map and previous hidden state are convolved with the attention kernel parameters $\mathbf{W}_a$ and $\mathbf{U}_a$ respectively. The corresponding bias unit is denoted by $b_a$. The ALSTM is a simple extension of the convolutional LSTM by which the input image $\mathbf{s}^{\langle t\rangle}$ is repeatedly propagated to the recurrent model and multiplied with $\text{softmax}(\mathbf{q}^{\langle t-1\rangle})$. Abawi~et~al.~\cite{abawi2021gasp} modify the ALSTM by restoring its sequential input property and attending to each element in the input sequence:
\begin{equation}
    \label{eq:alstm_atten}
    \mathbf{s}^{\langle t\rangle} = \text{softmax}(\mathbf{q}^{\langle t-1\rangle}) \odot \mathbf{s'}^{\langle t\rangle},
\end{equation}
where $\mathbf{s'}^{\langle t\rangle}$ represents the input map of the current timestep before applying attention. To integrate the modalities, we use the convolutional gating mechanism introduced by Arevalo~et~al.~(GMU)~\cite{arevalo2020gated}:
\begin{equation}
    \label{eq:gmu}
    \begin{aligned}
    \mathbf{j}^{(k)\langle t\rangle} &= \tanh{(\mathbf{W}_j^{(k)} * \mathbf{s}^{(k)\langle t\rangle} + b_j^{(k)})}, \\
    \mathbf{d}^{(k)\langle t\rangle} &= \sigma{(\mathbf{W}_d^{(k)} * [\mathbf{s}^{(1)\langle t\rangle}, ..., \mathbf{s}^{(K)\langle t\rangle}] + b_d^{(k)})}, \\
    \mathbf{j}^{\langle t\rangle} &= \sum_{k=1}^{K} \mathbf{d}^{(k)\langle t\rangle} \odot \mathbf{j}^{(k)\langle t\rangle}.
    \end{aligned}
\end{equation}
Here, $\mathbf{j}^{(k)\langle t\rangle}$ and $\mathbf{d}^{(k)\langle t\rangle}$ represent the gated projections for modality $k$ of all modalities $K$ at timestep $t$, along with their respective kernel parameters $\mathbf{W}_j^{(k)}$ and $\mathbf{W}_d^{(k)}$. The corresponding bias units are denoted by $b_j^{(k)}$ and $ b_d^{(k)}$. The output $\mathbf{j}^{\langle t\rangle}$ represents the final feature map of the gating module resulting from the Hadamard product of all modality-specific projections $\mathbf{j}^{(k)\langle t\rangle}$ and $\mathbf{d}^{(k)\langle t\rangle}$. The modality inputs \mbox{\{$\mathbf{s}^{(1)\langle t\rangle}$,~\dots~, $\mathbf{s}^{(K)\langle t\rangle}$\}} at timestep $t$ are concatenated across the channels as signified by the $[\mathbf{\cdot}{,}\mathbf{\cdot}]$ operator, and convolved with $\mathbf{W}_d^{(k)}$. We follow the integration paradigms introduced in GASP~\citep{abawi2021gasp} and adapt the sequential ALSTM to operate in conjunction with the GMU. One such integration paradigm entails performing the sequential gating followed by modality gating. This model is referred to as the Attentive Recurrent Gated Multimodal Unit (ARGMU), illustrated in~\subautoref[a]{fig:model_variants}. Alternatively, performing the gated integration after concatenating the input channels and propagating them to the sequential ALSTM is illustrated in~\subautoref[b]{fig:model_variants} and denoted by the Late ARGMU (LARGMU) variant. We describe ARGMU as a \textit{fusion} model, since feature integration occurs on modality-specific representations. LARGMU is a \textit{late integration} model, since the modality representations are concatenated before integrating them into a single representation.

\section{Experimental Setup}
\textcolor{black}{In this section, we describe the components of our experimental pipeline for conducting model training and evaluation. We also present the datasets used for training our models and the hyperparameter values chosen for those models. In contrast to the GASP model, we do not consider the saliency metrics as part of the loss function, we train each sample on the fixations of observers independently, and the training procedure as well as hyperparameters are adjusted to accommodate training on different datasets and task.}

\subsection{Evaluation Metrics}
Common scanpath prediction metrics measure the proximity of human fixation trajectories to those generated by the model~\citep{cristino2010scanmatch,dewhurst2012depends}. One shortfall of such approaches is the requirement to temporally align the scanpaths under comparison. This, however, adds a layer of complexity to streamed dynamic stimuli which could potentially cause scanpaths to diverge over time. \textcolor{black}{Moreover, scanpath metrics could potentially score arbitrary predictions as more accurate than scanpath generating models~\citep{kadner2024fixationprediction,kuemmerer2021scanpath}. We, therefore, follow the approach detailed by K\"ummerer~and~Bethge~\cite{kuemmerer2021scanpath} to evaluate our models}. Each sequence of input features, along with fixation histories, is used to generate a priority map for a single timestep. These maps are fed recursively to the model by appending them to the fixation history. One advantage to approaching scanpath evaluation as such is that it enables the usage of common metrics used to validate dynamic saliency models. Saliency metrics are generally more robust to incorrect predictions and do not require as many parametric assumptions as is the case with scanpath metrics~\citep{kuemmerer2021scanpath}, for instance, ScanMatch~\citep{cristino2010scanmatch} and MultiMatch~\citep{dewhurst2012depends}. Moreover, our models do not generate sequences, instead they predict the last fixation map---blurred fixation point---which is conditioned on previous ground-truth maps in the fixation history. Comparing the entire sequence using scanpath metrics would not accurately represent the performance of our models, since they predict single fixation maps at each timestep instead of the entire scanpaths. We therefore use the following saliency metrics to evaluate our model predictions:

\noindent \textbf{Normalized Scanpath Saliency (NSS)} is a location-based metric~\cite{bylinskii2019eval} to measure the correspondence between ground-truth and predicted attention maps according to fixated locations. False positives have an effect on the NSS score, making it suitable for quantifying the quality of noisy predictions. A high positive NSS score indicates that the model accurately predicts the locations of fixations as expressed by:
\begin{equation}
    \label{eq:metric_nss}
     \begin{aligned}
    {\!N\!S\!S} &= {\sum\limits_{x,y}} \frac{\mathbf{\hat{m}}_{x,y} - \mu(\mathbf{\hat{m}}_{x,y})}{\rho(\mathbf{\hat{m}}_{x,y})} \odot \mathbf{m}_{<x,y>},
    \end{aligned}
\end{equation}
where $\mathbf{m}_{<x,y>}$ refers to the ground-truth fixation point rather than the continuous priority map expressed as $\mathbf{m}_{x,y}$. The mean of the priority map is denoted by $\mu$, whereas the standard deviation by $\rho$. Our model predicts the probability of fixations, which is not restricted to a single point in space. Having multiple predicted fixations is the desired outcome in our scanpath modeling approach since a single prediction is an unrealistic assumption and would imply that our model is not robust, i.e., humans do not always look toward the same point when shown a sequence of images multiple times, therefore, having a definite fixation prediction indicates overfitting. Multiple fixations result in a lower NSS score for the priority map of an individual as compared to the group attention map. We, therefore, rely on the NSS rather as an indicator of the relative difference between the individual scanpath prediction models. 

\noindent \textbf{Area Under the ROC-Curve (AUC)} is another location-based metric, which classifies whether a pixel in space is fixated or not. We rely on an AUC variant developed by Judd~et~al.~\cite{judd2009learning}, denoted hereafter by \textbf{AUCJ}. The advantage of using AUC as a measure of quality for our task is that the true and false positive rates are functions of the number of fixated and unfixated pixels respectively. Since we have a single fixated ground-truth pixel for an individual, the weighing of true positives avoids skewing our evaluation toward false examples, providing a clear interpretation of our model's performance. 

\subsection{Datasets}
For our study, we used two existing datasets comprising eye-gaze data from observers viewing conversational videos. \textcolor{black}{The choice of dataset is governed by two requirements. First, each observer's identifier must be identical across videos. This allows us to extract the fixation history and ground-truth priority maps to train our personalized scanpath prediction models. Second, given our reliance on the GASP model, which is designed specifically for enhancing saliency prediction on social videos, we require a dataset that allows for the extraction of social cues. This dictates that all videos should include humans with visible faces. The datasets chosen fulfill these requirements.}

These datasets, sourced from YouTube and Youku, feature social videos with gaze data from a nearly identical observer count, ensuring consistency in our comparisons between the datasets. We note that each observer watched all the videos within a given dataset, enabling our model to distinguish prototypical gaze behaviors for predicting scanpaths.
A dataset by Xu~et~al.~\cite{xu2018findwho} consists of 65 conversational videos. Gaze data were collected from 39 observers, including 26 males and 13 females, between the ages of 20 and 49. Hereafter, we refer to this dataset as \mbox{\textit{FindWho}~\citep{xu2018findwho}}.
The \textit{MVVA}~\citep{liu2020mvva, qiao2023mvva} dataset, on the other hand, is more extensive with 300 conversational videos. Gaze data were derived from 34 observers, including 21 males and 13 females, between the ages of 20 and 54. In all analyses involving the MVVA dataset, one observer was excluded due to noisy data, reducing the total number of observers to 33. An overview of the FindWho and MVVA dataset properties is shown in~\autoref{tab:datasets}. 
\begin{table}[!btp]
\caption{Experimental setup and dataset properties.\label{tab:datasets}}
\centering
\begin{adjustbox}{width=0.44\textwidth}
\small
\begin{tabular}{c|c|c}
\hline
\textbf{Property} & MVVA~\citep{liu2020mvva} & FindWho~\citep{xu2018findwho}\\
\hline
Distance to monitor & \around55~cm & \around60~cm\\
Monitor resolution & 1280\cross720~px & 1280\cross720~px \\
 & (16:9) & (16:9)\\
Monitor size & 23-inch & 23.8-inch\\
Video duration & 10-30s & \around20s\\
Frames per second & 30 & 25\\
Audio channels & Stereo & Monaural\\
Head-pose & Free & Fixed\\
\hline
No. training videos & 210 (70\%) & 46 (70\%)\\
No. validation videos & 30 (10\%) & -\\
No. test videos & 60 (20\%) & 19 (30\%)\\
No. observers & 34 (1 excl.) & 39\\
\hline
\end{tabular}
\end{adjustbox}
\end{table}

\subsection{Model Training and Evaluation}
The individual and unified scanpath prediction models are trained on an NVIDIA GeForce GTX 3080 Ti GPU with 12~GB VRAM and 32~GB RAM. The individual model training process requires separate models for each observer. This is a highly demanding procedure, necessitating the distribution of models across multiple machines and GPUs. We orchestrate these processes through a custom workflow manager, developed using the Wrapyfi~\citep{wrapyfi2024abawi} framework, allowing us to exchange completion logs across training instances over message-oriented middleware.
All social cue detectors and models are adopted from GASP~\citep{abawi2021gasp} and are implemented in PyTorch~\citep{pytorch2019paszke}. 

The two model architecture variants, DAM + ARGMU (context size $T'=8$) and DAM + LARGMU (context size $T'=10$) are initialized with their GASP parameters, trained on the social subset of the AVE~\citep{tavakoli2020deep} dataset. We fine-tune the individual and unified models on the MVVA and FindWho datasets separately, for 10 and 50
 epochs respectively. We used early stopping with $\delta_{min}=.0001$ and a patience of $3$. This resulted in all models and architectures converging on average at epoch 6 for MVVA and epoch 11 for FindWho.

The individual models have a predefined observer set for all samples, whereas the unified model randomly selects an observer for each training sample. During training, frame samples overlap by 90\%. For the late integration architecture, this resulted in a training time of 90 minutes per epoch on the MVVA dataset and 20 minutes on the FindWho dataset. For the early fusion architecture, the training time per epoch is 76 minutes on the MVVA dataset and 17 minutes on the FindWho dataset. The time required to train a single individual and unified model is identical given the same architecture and dataset. However, since individual models are trained separately for each observer, this results in 33~\cross~epochs for the MVVA dataset and 39~\cross~epochs for the FindWho dataset, in comparison to the unified models.

The batch size is set to $48$ with gradient accumulation over $4$ mini-batches, where each batch element contains the entire sequence of modality representations to match the context size of any given trained model. Each model is trained using the Adam optimizer, setting $\beta_1=.9$, $\beta_2=.999$, and the learning rate $\alpha=.001$. 

Models are evaluated on subsampled video frames at 10 fps, with no overlap between consecutive frames. Evaluation is performed on the basis of one-step-ahead prediction unless specified otherwise. All observer predictions are evaluated independently for both individual and unified models. All unified and individual models were trained and evaluated over 5 trials. 

\subsection{Saliency Losses}
\label{sec:salloss}
In order to train the model, we employ the loss functions introduced by Cornia et al.~\cite{cornia2018predicting}.
The loss functions are weighted, summed, and applied to the final layer, implying that the learnable parameters of our model, specifically the modality encoder and fusion model parameters are optimized. We denote the overall loss function by $\mathcal{L}_{P\!F\!D\!M}$ and define it as:

\begin{equation}
    \label{eq:sum_loss}
    \mathcal{L}_{\!P\!F\!D\!M} = \mathcal{L}_{\!N\!L\!L} + \mathcal{L}_{\!K\!L\!D},
\end{equation}
where $\mathcal{L}_{\!N\!L\!L}$ computes the \textit{negative log-likelihood} loss as expressed by Sun~et~al.~\cite{sun2019visual} between the ground-truth and predicted priority maps, followed by minimization of the \textit{Kullback-Leibler divergence}.
\begin{equation}
    \label{eq:loss_1}
     \begin{aligned}
    \mathcal{L}_{\!N\!L\!L} &= - \lambda_{\!N\!L\!L} \cdot \sum\limits_{x,y}  \mathbf{m}_{<x,y>} \odot \log(\mathbf{\hat{m}}_{<x,y>}) \\
    &~+ (1 -  \mathbf{m}_{<x,y>}) \odot (1 - \log(\mathbf{\hat{m}}_{<x,y>})), \\
    \mathcal{L}_{\!K\!L\!D}^{+} &= \sum\limits_{x,y}  \mathbf{m}_{x,y} \odot (\log(\mathbf{m}_{x,y}) - \log(\mathbf{\hat{m}}_{x,y})), \\
    \mathcal{L}_{\!K\!L\!D}^{-} &= \sum\limits_{x,y} (1 - \mathbf{m}_{x,y}) \odot ((1 - \log(\mathbf{m}_{x,y})) \\
    &~- (1  - \log(\mathbf{\hat{m}}_{x,y}))), \\
    \mathcal{L}_{\!K\!L\!D} &= - \lambda_{\!K\!L\!D} \cdot (\mathcal{L}_{\!K\!L\!D}^{+} + \mathcal{L}_{\!K\!L\!D}^{-}).
    \end{aligned}
\end{equation}
Algorithmically, the cross-entropy loss utilized in GASP and $\mathcal{L}_{\!N\!L\!L}$ are identical, however, $\mathcal{L}_{\!N\!L\!L}$ operates on the fixation point, replacing the priority map $\mathbf{m}_{x,y}$ with $\mathbf{m}_{<x,y>}$. Without the negative log-likelihood loss, the models require more epochs (3 to 7 additional epochs) to converge to similar states relying purely on $\mathcal{L}_{\!K\!L\!D}$. 

The inverted stream of our DAM layer has a separate output head for each timestep. We compute the cross-entropy between the DAM prediction and ground-truth fixation density maps for all pixels summed over all timesteps. The $\mathcal{L}_{DAM}$ is computed to optimize the inverted stream parameters. These parameters are transferred to the tied direct stream. The direct stream parameters are frozen throughout the training phase. In this manner, we are able to emphasize weaker modalities, intensifying the propagation of noisy signals to the sequential integration model, effectively acting as a regularizer.

We employed the Tree-structured Parzen Estimator (TPE) method using Hyperopt~\cite{hyperopt2013bergstra} for hyperparameter optimization, to identify the optimal loss weights.\footnote{All search trials were applied to the late integration variant (DAM + LARGMU, $T'=10$) with encoders pretrained on the AVE~\citep{tavakoli2020deep} dataset---excluding the fixation history module---and fine-tuned for 6 epochs on the MVVA~\citep{liu2020mvva} dataset. The TPE minimized the validation loss on MVVA.} The considered weight range was \mbox{$\in~[.01, 1]$} sampled from a log-normal distribution. Based on the TPE's results after $90$ trials, the loss weight for $\mathcal{L}_{\!K\!L\!D}$ is set to
 $\lambda_{\!K\!L\!D}=.94$, whereas the loss weight for $\mathcal{L}_{\!N\!L\!L}$ is determined to be $\lambda_{\!N\!L\!L}=.03$. For $\mathcal{L}_{\!D\!A\!M}$, the loss weight is established as $\lambda_{\!D\!A\!M}=.61$. 

\section{Results}

We evaluated our late integration and early fusion architectures on the FindWho and MVVA datasets. This assessment was conducted by comparing each individual model's prediction against the last fixation in the individual observer's scanpath (\textit{1~vs~1}) and against the group---all observers---fixation density map, excluding the individual's data (\textit{1~vs~infinity}). Moreover, we conducted statistical analyses to compare the unified and individual models. We then performed a social cue ablation study on the two unified model variants: late integration and early fusion. Finally, we tested the unified models to quantify the degradation of predictions over longer horizons beyond the next fixation point as detailed in~\autoref{sec:fix_hist} under multi-step-ahead evaluation. The mean values of the metric scores represent the performance of our models independently across all evaluation videos for every observer. Trial mean values are reported for all experiment results unless stated otherwise.

\subsection{Individual Models}
\label{sec:results_architecture_dataset_1v1_1vinf}
\begin{figure*}[!ht]
\centering
\subcaptionbox*{}{\includegraphics[width=0.11\textwidth, trim = 23em 15em 0em 3em, clip=true]{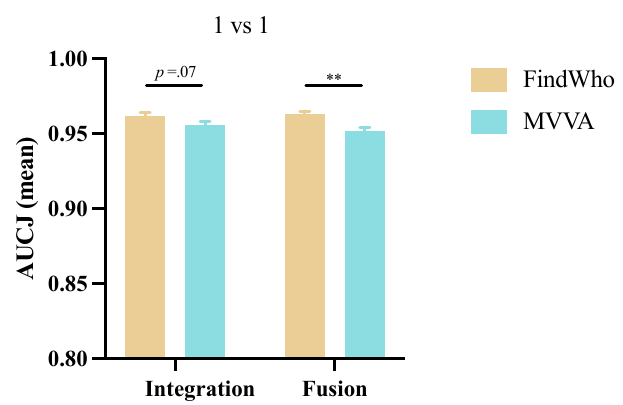}}%
\subcaptionbox*{}{\includegraphics[width=0.11\textwidth, trim = 23em 12.9em 0em 5.36em, clip=true]{src/exp/architecture_dataset_1vs1_1vsInfinite_aucj_nss_mean_std/Aucj_1_vs_1_mean.pdf}}%
\vfill
\subcaptionbox{1~vs~1}{\includegraphics[width=0.25\textwidth, trim = 0 0 8em 2em, clip=true]{src/exp/architecture_dataset_1vs1_1vsInfinite_aucj_nss_mean_std/Aucj_1_vs_1_mean.pdf}}%
\hfil
\subcaptionbox{1~vs~infinity}{\includegraphics[width=0.25\textwidth, trim = 0 0 8em 2em, clip=true]{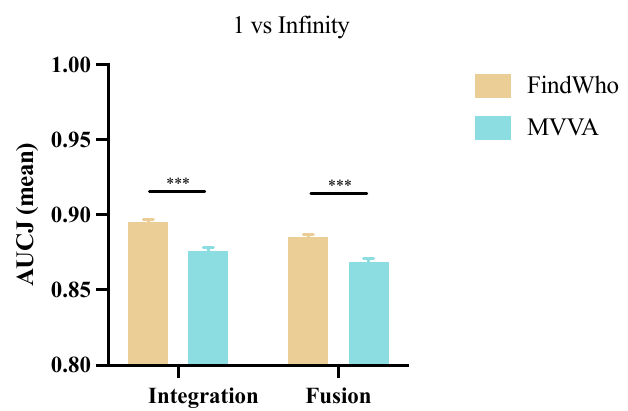}}%
\hfil
\subcaptionbox{1~vs~1}{\includegraphics[width=0.25\textwidth, trim = 0 0 8em 2em, clip=true]{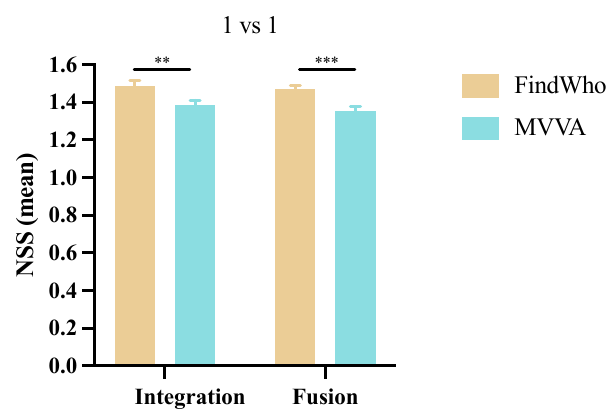}}%
\hfil
\subcaptionbox{1~vs~infinity}{\includegraphics[width=0.25\textwidth, trim = 0 0 8em 2em, clip=true]{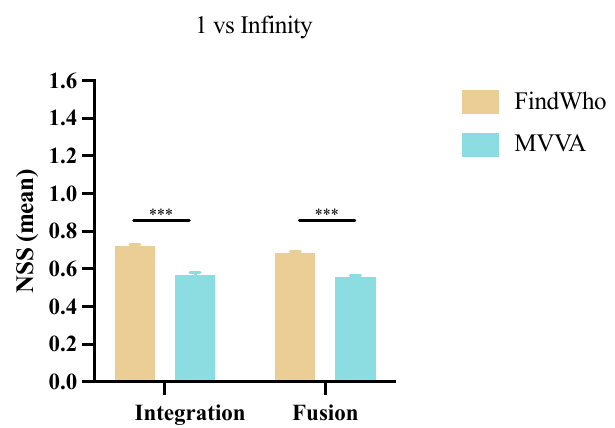}}%

\caption{The individual model \textit{1~vs~1} and \textit{1~vs~infinity} evaluations on the FindWho~\citep{xu2018findwho} and MVVA~\citep{liu2020mvva} datasets, across the two GASP~\citep{abawi2021gasp} variants extended with fixation history modules. \textbf{(a,d)} visualize the mean values of the scores across all samples.\\ \footnotesize{$*\!*$ denotes $.001 < p < .01$ and $*\!*\!*$ $ p < .001$}}
\label{fig:1vs1_1vsinfinity}
\end{figure*}

\begin{figure*}[!ht]
\centering
\subcaptionbox*{}{\includegraphics[width=0.11\textwidth, trim = 23em 15em 0em 3em, clip=true]{src/exp/architecture_dataset_1vs1_1vsInfinite_aucj_nss_mean_std/Aucj_1_vs_1_mean.pdf}}%
\subcaptionbox*{}{\includegraphics[width=0.11\textwidth, trim = 23em 12.9em 0em 5.36em, clip=true]{src/exp/architecture_dataset_1vs1_1vsInfinite_aucj_nss_mean_std/Aucj_1_vs_1_mean.pdf}}%
\vfill

\subcaptionbox{1~vs~1 (variance across videos)}{\includegraphics[width=0.24\textwidth, trim = 0 0 8em 2em, clip=true]{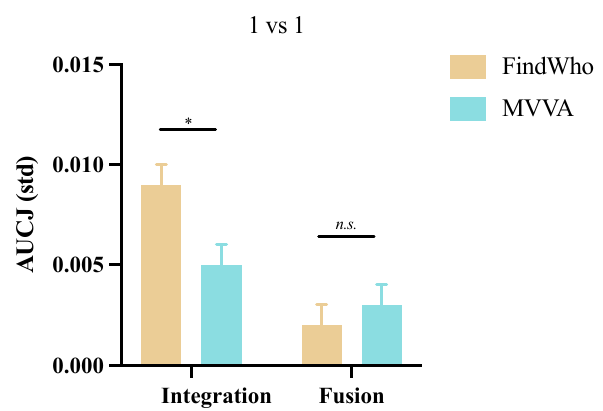}}%
\hfil
\subcaptionbox{1~vs~infinity (variance across videos)}{\includegraphics[width=0.24\textwidth, trim= 0 0 8em 2em, clip=true]{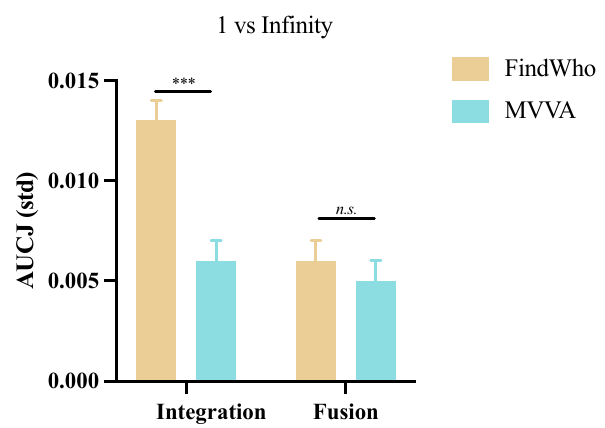}}%
\hfil
\subcaptionbox{1~vs~1 (variance across videos)}{\includegraphics[width=0.235\textwidth, trim = 0 0 8em 2em, clip=true]{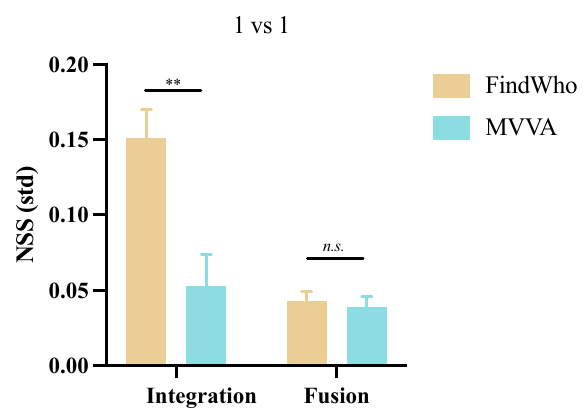}}%
\hfil
\subcaptionbox{1~vs~infinity (variance across videos)}{\includegraphics[width=0.235\textwidth, trim= 0 0 8em 2em, clip=true]{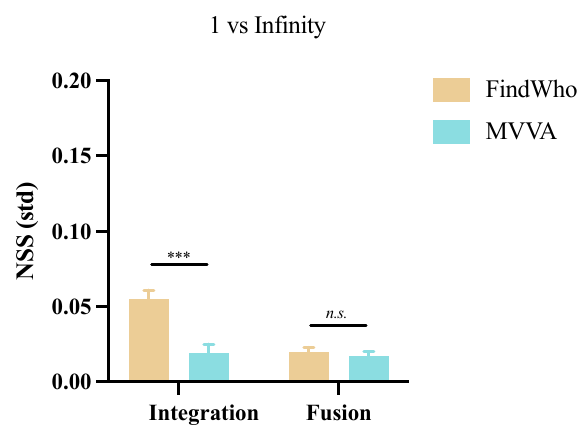}}%
\caption{The individual model \textit{1~vs~1} and \textit{1~vs~infinity} evaluations on the FindWho~\citep{xu2018findwho} and MVVA~\citep{liu2020mvva} datasets, across the two GASP~\citep{abawi2021gasp} variants extended with fixation history modules. \textbf{(a,d)} visualize the standard deviation of the scores across all testing videos per individual observer.\\ \footnotesize{$*$ denotes $.01 < p < .05$, $*\!*$ $.001 < p < .01$, $*\!*\!*$ $ p < .001$, and \textit{n.s.} denotes no significance.}}
\label{fig:1vs1_1vsinfinity_vid_var}
\end{figure*}
To examine the impact of the model architecture, dataset size, and their interaction effects on the models' performances, a 2 (integration~vs~fusion)\cross2 (FindWho~vs~MVVA) mixed analysis of variance (ANOVA) was conducted. Specifically, the model architecture was a within-subject factor, and the dataset size was a between-subject factor. The models' performances were measured by AUCJ (mean and std) and NSS (mean and std). All metrics were measured between observers, except for `variance across videos' experiments, where the metrics measure the variance in video results per observer---the score variances across the videos averaged for all observers.

\subsubsection{1~vs~1 Evaluations}

Significant main effects and interactions were observed in terms of the AUCJ and NSS metric scores. For the performance of the datasets, the smaller FindWho dataset outperformed the larger MVVA dataset, with significant differences (AUCJ: $F(1, 68) = 7.04$, $p < .05$, $\eta_p^2 = .09$ as shown in~\subautoref[a]{fig:1vs1_1vsinfinity}; NSS: ($F(1, 68) = 11.79$, $p < .01$, $\eta_p^2 = .14$) as shown in~\subautoref[c]{fig:1vs1_1vsinfinity}). The interaction effect between model architecture and dataset size was significant in terms of the AUCJ score ($F(1, 68) = 11.08$, $p < .01$, $\eta_p^2 = .14$), where the integration architecture significantly outperformed the fusion architecture on evaluating the MVVA dataset ($p < .01$). However, no significant differences were found on evaluating the FindWho dataset ($p = .19$). The NSS metric, however, did not show a significant interaction between architecture and dataset size ($F(1, 68) = .19$, $p = .67$, $\eta_p^2 = .003$).

Variance across videos indicates instability in the model predictions. For the AUCJ score, the integration architecture resulted in a significantly larger variance compared to the fusion architecture ($F(1, 68) = 20.94$, $p < .01$, $\eta_p^2 = .23$). The MVVA dataset performed significantly better---lower variance---within the integration architecture ($p < .05$) and no significant performance difference within the fusion architecture ($p = .62$) in terms of the AUCJ score as shown in~\subautoref[a]{fig:1vs1_1vsinfinity_vid_var}. The interaction effect between architecture and dataset size was also significant in terms of the AUCJ score ($F(1, 68) = 6.44$, $p < .05$, $\eta_p^2 = .08$). Similarly, for the NSS score, the fusion architecture outperformed the integration architecture ($F(1, 68) = 17.96$, $p < .001$, $\eta_p^2 = .20$), and a significant interaction between architecture and dataset size was observed ($F(1, 68) = 10.78$, $p < .01$, $\eta_p^2 = .13$). The MVVA dataset again performed significantly better within the integration architecture ($p < .05$) but no significant differences were observed within the fusion architecture in terms of the NSS score as shown in~\subautoref[c]{fig:1vs1_1vsinfinity_vid_var}.

\subsubsection{1~vs~infinity Evaluations}

In terms of the AUCJ and NSS metric scores, significant main effects were observed for both architecture and dataset size, with no significant interaction effects in some cases. For AUCJ mean values, the integration architecture demonstrated better performance compared to the fusion architecture ($F(1, 68) = 36.91$, $p < .001$, $\eta_p^2 = .35$). The FindWho dataset outperformed the MVVA dataset in terms of the AUCJ score ($F(1, 68) = 40.98$, $p < .001$, $\eta_p^2 = .37$) as shown in~\subautoref[b]{fig:1vs1_1vsinfinity}. However, the interaction between architecture and dataset size was not significant in terms of the AUCJ score ($F(1, 68) = 1.32$, $p = .26$, $\eta_p^2 = .02$). Similar trends were observed in terms of the NSS score, where the integration architecture also resulted in a better performance compared to the fusion architecture ($F(1, 68) = 24.35$, $p < .001$, $\eta_p^2 = .26$). The FindWho dataset significantly outperformed the MVVA dataset in terms of the NSS score ($F(1, 68) = 82.88$, $p < .001$, $\eta_p^2 = .54$) as shown in~\subautoref[d]{fig:1vs1_1vsinfinity}, due to the smaller size of the former's test set. A significant interaction effect between architecture and dataset size was observed in terms of the NSS score ($F(1, 68) = 6.79$, $p < .05$, $\eta_p^2 = .088$). More specifically, the FindWho dataset within the integration architecture outperformed the fusion architecture ($p < .001$), with no significant difference observed for the MVVA dataset ($p = .12$).

Unlike \textit{1~vs~1} variance across videos, a higher variance in \textit{1~vs~infinity} is interpreted as a positive outcome, since it indicates that the model predicts scanpaths that are personalized to the individual observer. Significant effects  were observed for the architecture and dataset size in terms of the AUCJ and NSS metric scores. In terms of the AUCJ score, the integration architecture outperformed the fusion architecture ($F(1, 68) = 27.10$, $p < .001$, $\eta_p^2 = .28$). The FindWho dataset performed better than the MVVA dataset in terms of the AUCJ score ($F(1, 68) = 9.94$, $p < .01$, $\eta_p^2 = .12$) as shown in~\subautoref[b]{fig:1vs1_1vsinfinity_vid_var}. A significant interaction effect was also observed, particularly within the integration architecture, where the FindWho dataset significantly outperformed MVVA dataset in terms of the AUCJ score ($p < .001$). Similar significant main effects were observed for both architecture ($F(1, 68) = 15.36$, $p < .001$, $\eta_p^2 = .18$) and dataset size ($F(1, 68) = 16.59$, $p < .001$, $\eta_p^2 = .19$) in terms of the NSS score. The FindWho dataset within the integration architecture also performed significantly worse than the MVVA dataset in terms of the NSS score ($F(1, 68) = 12.36$, $p < .01$, $\eta_p^2 = .15$) as shown in~\subautoref[d]{fig:1vs1_1vsinfinity_vid_var}. However, no significant differences observed within the fusion architecture.

\begin{figure*}[!htbp]
\centering
\subcaptionbox{Integration}{\includegraphics[width=0.25\textwidth, trim = 0 0 10em 2em, clip=true]{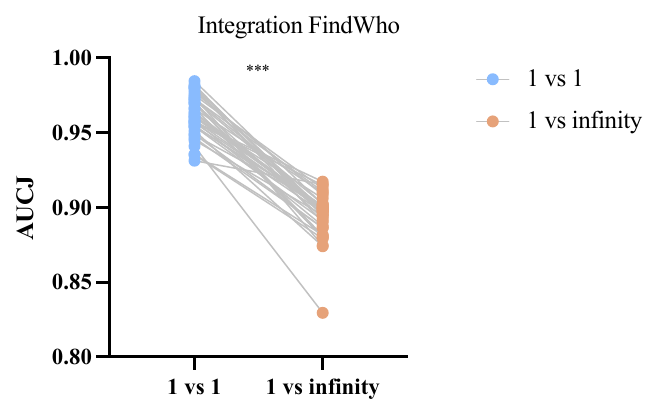}}%
\hfill
\subcaptionbox{Fusion}{\includegraphics[width=0.242\textwidth, trim = 0 0 10em 2em, clip=true]{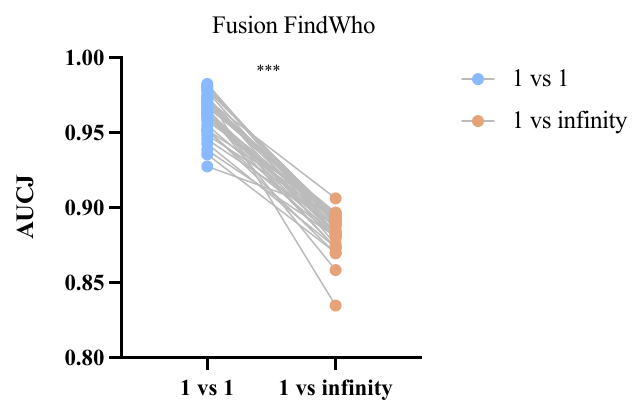}}%
\hfill
\subcaptionbox{Integration}{\includegraphics[width=0.222\textwidth, trim = 0 0 9.5em 2em, clip=true]{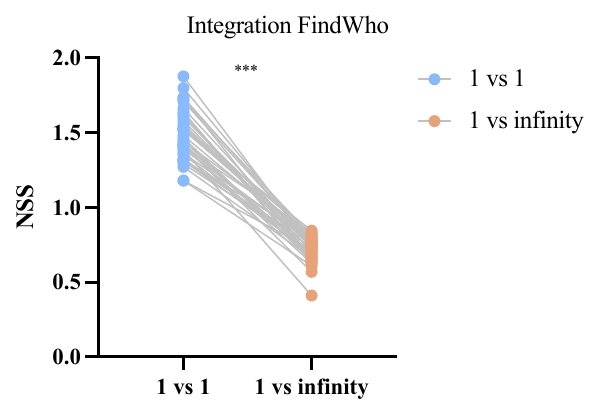}}%
\hfill
\subcaptionbox{Fusion}{\includegraphics[width=0.246\textwidth, trim= 0 0 10em 2em, clip=true]{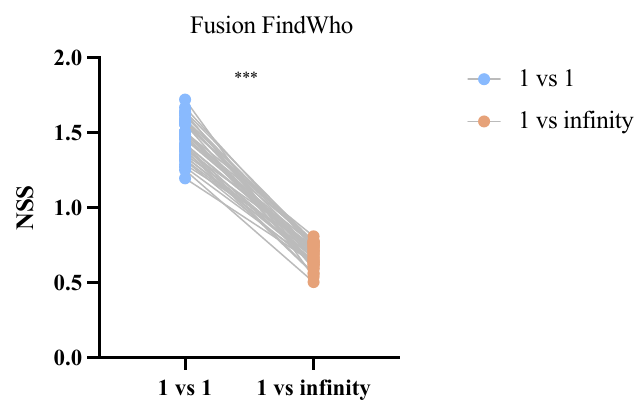}}%
\caption{The individual model \textit{1~vs~1} and \textit{1~vs~infinity} evaluations on the FindWho~\citep{xu2018findwho} dataset, across the two GASP~\citep{abawi2021gasp} variants extended with fixation history modules. The AUCJ scores are measured for the \textbf{(a)} integration and \textbf{(b)} fusion architectures, as well as NSS scores for the \textbf{(c)} integration and \textbf{(d)} fusion architectures.\\ \footnotesize{$*\!*\!*$ denotes $ p < .001$.}}
\label{fig:findwho_dots_t_test_1vsinfinity}
\end{figure*}

\begin{figure*}[!htbp]
\subcaptionbox{Integration}{\includegraphics[width=0.25\textwidth, trim = 0 0 10em 2em, clip=true]{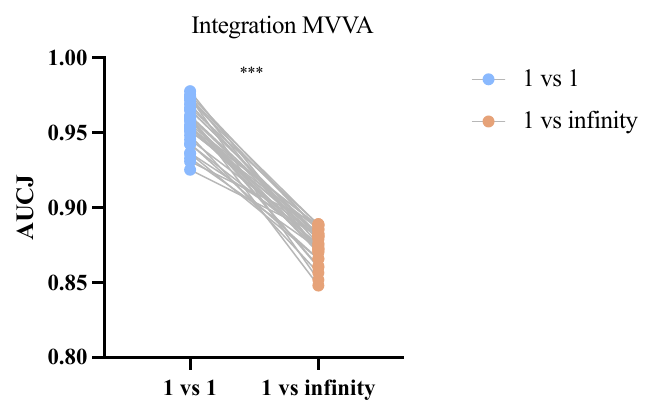}}%
\hfill
\subcaptionbox{Fusion}{\includegraphics[width=0.25\textwidth, trim = 0 0 10em 2em, clip=true]{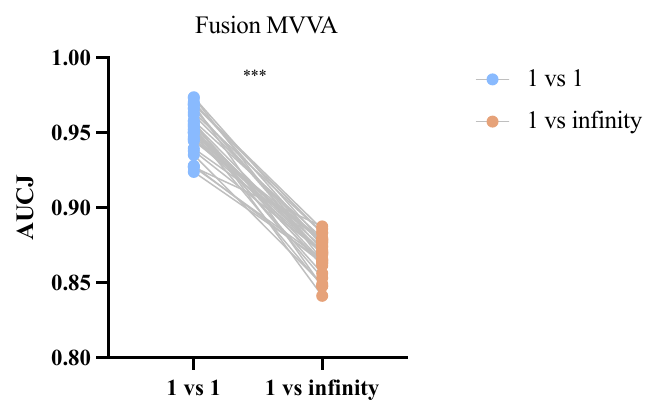}}%
\hfill
\subcaptionbox{Integration}{\includegraphics[width=0.245\textwidth, trim = 0 0 10em 2em, clip=true]{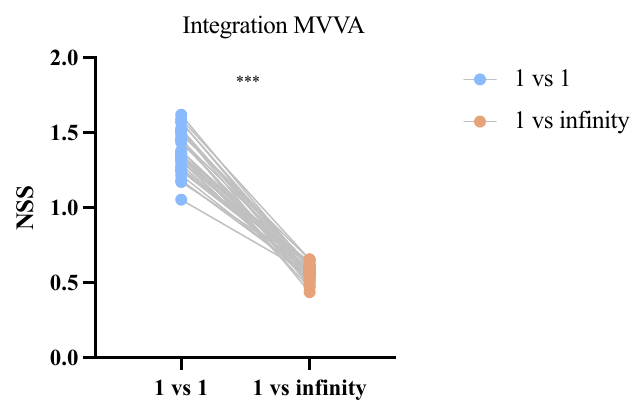}}%
\hfill
\subcaptionbox{Fusion}{\includegraphics[width=0.246\textwidth, trim= 0 0 10em 2em, clip=true]{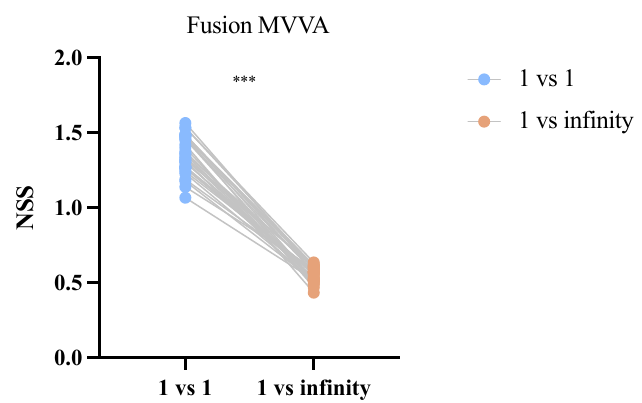}}%
\caption{The individual model \textit{1~vs~1} and \textit{1~vs~infinity} evaluations on the MVVA~\citep{liu2020mvva} dataset, across the two GASP~\citep{abawi2021gasp} variants extended with fixation history modules. The AUCJ scores are measured for the \textbf{(a)} integration and \textbf{(b)} fusion architectures, as well as NSS scores for \textbf{(c)} the integration and \textbf{(d)} fusion architectures.\\ \footnotesize{$*\!*\!*$ denotes $ p < .001$.}}
\label{fig:mvva_dots_t_test_1vsinfinity}
\end{figure*}
\subsubsection{Comparison between 1~vs~1 and 1~vs~infinity Evaluations}
A group of paired-samples t-tests showed that for both architectures, performance in the \textit{1~vs~1} evaluation was significantly better than in \textit{1~vs~infinity}, in terms of AUCJ and NSS metric scores, $ps < .001$. Details of the analyses on the FindWho and MVVA datasets can be found in \autoref{tab:t-test_FindWho} and \autoref{tab:t-test_MVVA} respectively. The results are shown in \autoref{fig:findwho_dots_t_test_1vsinfinity} and \autoref{fig:mvva_dots_t_test_1vsinfinity}.

\begin{table}[hbtp]
    \centering
    \caption{Individual models trained and evaluated on the FindWho~\citep{xu2018findwho} dataset with \textit{1~vs~1} and \textit{1~vs~infinity} comparisons in terms of AUCJ and NSS scores. The t-test degrees of freedom are shown within parentheses.}
    \resizebox{0.40\textwidth}{!}{%
    \begin{tabular}{l|cc|cc}
        \hline
        \multirow{2}{*}{} & \multicolumn{2}{c}{\textbf{Integration}} & \multicolumn{2}{c}{\textbf{Fusion}}\\
        & AUCJ$\uparrow$ & NSS$\uparrow$ & AUCJ$\uparrow$ & NSS$\uparrow$ \\
        \hline
        \textit{1~vs~1}& 0.962 & 1.488 & 0.963 & 1.467\\
        \textit{1~vs~infinity}& 0.895 & 0.719 & 0.885 & 0.682\\
        t-value (df = 38)& 23.23 & 27.89 & 27.45 & 36.90 \\
        \hline
    \end{tabular}
    }
    \label{tab:t-test_FindWho}
\end{table}

\begin{table}[!btp]
    \centering
    \caption{Individual models trained and evaluated on the MVVA~\citep{liu2020mvva} dataset with \textit{1~vs~1} and \textit{1~vs~infinity} comparisons in terms of AUCJ and NSS scores. The t-test degrees of freedom are shown within parentheses.}
    \resizebox{0.40\textwidth}{!}{%
    \begin{tabular}{l|cc|cc}
        \hline
        \multirow{2}{*}{} & \multicolumn{2}{c}{\textbf{Integration}} & \multicolumn{2}{c}{\textbf{Fusion}}\\
        & AUCJ$\uparrow$ & NSS$\uparrow$ & AUCJ$\uparrow$ & NSS$\uparrow$ \\
        \hline
        \textit{1~vs~1}& 0.956 & 1.383 & 0.952 & 1.353\\
        \textit{1~vs~infinity}& 0.876 & 0.567 & 0.869 & 0.556\\
        t-value (df = 32)& 25.54 & 32.57 & 27.41 & 36.42\\
        \hline
    \end{tabular}
    }
    \label{tab:t-test_MVVA}
\end{table}

\subsection{Unified~vs~Individual Models}
\label{sec:results_interaction_effects}
\begin{figure*}[ht]
\centering
\subcaptionbox*{}{\includegraphics[width=0.16\textwidth, trim = 22em 15em 0em 3em, clip=true]{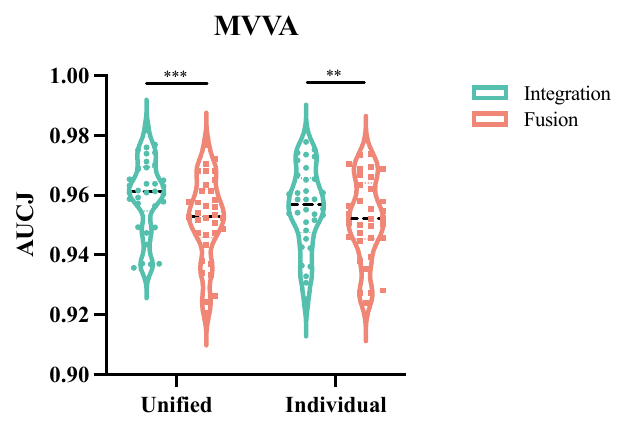}}%
\subcaptionbox*{}{\includegraphics[width=0.16\textwidth, trim = 22em 13.74em 0em 5.36em, clip=true]{src/exp/results_model_architecture_dataset_new/AUCJ_MVVA_model_architecture.pdf}}%
\vfill
\subcaptionbox{MVVA}
{\includegraphics[width=0.25\textwidth, trim = 0 0 10em 2.5em, clip=true]{src/exp/results_model_architecture_dataset_new/AUCJ_MVVA_model_architecture.pdf}}%
\hfill
\subcaptionbox{FindWho}{\includegraphics[width=0.248\textwidth, trim = 0 0 10em 2.5em, clip=true]{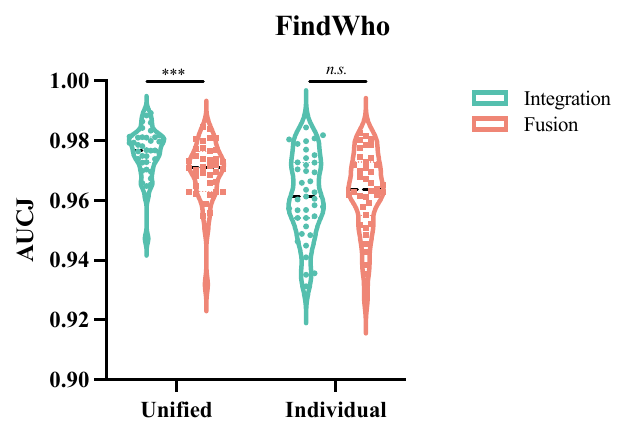}}%
\hfill
\subcaptionbox{MVVA}{\includegraphics[width=0.24\textwidth, trim = 0 0 10em 2.5em, clip=true]{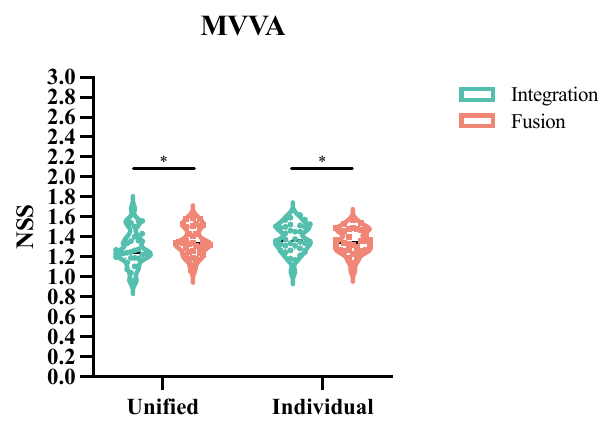}}%
\hfill
\subcaptionbox{FindWho}{\includegraphics[width=0.24\textwidth, trim= 0 0 10em 2.5em, clip=true]{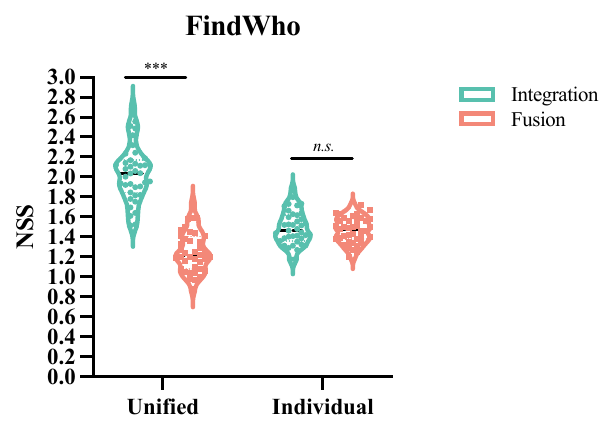}}%
\caption{The unified and individual model comparisons across the two GASP~\citep{abawi2021gasp} variants extended with fixation history modules. The AUCJ scores are measured when evaluating the models on the \textbf{(a)} MVVA~\citep{liu2020mvva} and \textbf{(b)} FindWho~\citep{xu2018findwho} datasets. The NSS scores are also measured for the \textbf{(c)} MVVA and \textbf{(d)} FindWho datasets.\\ \footnotesize{$*$ denotes $.01 < p < .05$, $*\!*$ $.001 < p < .01$, $*\!*\!*$ $ p < .001$, and \textit{n.s.} denotes no significance.}
\label{fig:effects_models_modelarchs_datasets}}
\end{figure*}
To examine the impact of the model, model architecture, dataset size, and their interaction effects on the models' performances, a 2 (unified model~vs~individual model)\cross2 (integration~vs~fusion)\cross2 (FindWho~vs~MVVA) mixed analysis of variance (ANOVA) was conducted. Specifically, the model and model architecture were within-subject factors, and the dataset size was a between-subject factor. 

For both the AUCJ and NSS metric scores, significant main effects were observed. The unified model demonstrated better performance over the individual models (AUCJ: unified $.964 \pm .001$ vs individual $.958 \pm .002$, $p < .001$; NSS: unified $1.480 \pm .022$ vs individual $1.424 \pm .016$, $p < .01$). The integration architecture significantly outperformed the fusion architecture (AUCJ: integration $.964 \pm .001$ vs fusion $.959 \pm .001$, $p < .001$ as shown in~\subautoref[a]{fig:effects_models_modelarchs_datasets} and~\subautoref[b]{fig:effects_models_modelarchs_datasets}; NSS: integration $1.548 \pm .020$ vs fusion $1.356 \pm .016$, $p < .001$ as shown in~\subautoref[c]{fig:effects_models_modelarchs_datasets} and~\subautoref[d]{fig:effects_models_modelarchs_datasets}). Additionally, the smaller FindWho dataset achieved better results compared to the larger MVVA dataset (AUCJ: FindWho $.968 \pm .002$ vs MVVA $.955 \pm .002$, $p < .001$; NSS: FindWho $1.561 \pm .023$ vs MVVA $1.344 \pm .024$, $p < .001$).

For two-factor interactions, we observed a significant interaction in the unified model which performed better with the integration architecture compared to the fusion architecture (AUCJ: unified with integration $.968 \pm .001$ vs fusion $.961 \pm .001$, $p < .001$; NSS: unified with integration $1.660 \pm .023$ vs fusion $1.302 \pm .020$, $p < .001$). The unified model significantly outperformed individual models when trained with the FindWho dataset (AUCJ: unified $.973 \pm .002$ vs individual $.963 \pm .002$, $p < .001$; NSS: unified $1.645 \pm .030$ vs individual $1.478 \pm .023$, $p < .001$). However, this difference diminished with the MVVA dataset. The integration architecture consistently yielded better results across both datasets, with particularly better performance on the FindWho dataset (AUCJ: integration $.969 \pm .002$ vs fusion $.966 \pm .002$, $p < .01$; NSS: integration $1.763 \pm .028$ vs fusion $1.360 \pm .023$, $p < .001$).

For three-factor interactions, we observed a significant interaction across the dataset, model architecture, and model ($p < .05$). The unified model performed better on the FindWho dataset regardless of the architecture used in terms of the AUCJ score (integration $.98 \pm .002$ vs fusion $.97 \pm .002$). The unified model performed significantly better with the integration architecture on the FindWho dataset in terms of the NSS score($2.035 \pm .038$ vs individual $1.488 \pm .025$, $p < .001$), but individual models performed better with the fusion architecture on the MVVA dataset.

\subsection{Social Cue Ablation}
\label{sec:results_socialcueablation}
\begin{table}[!hb]
    \centering
    \caption{The unified models of the two GASP~\citep{abawi2021gasp} variants extended with fixation history modules having social cue modalities ablated. The fusion and integration model architectures are trained and evaluated on the MVVA~\citep{liu2020mvva} and FindWho~\citep{xu2018findwho} datasets. The first combination group from the \textit{top} signifies models with a single cue modality, the second combination group signifies models with two cue modalities, and the final combination group signifies models with all cue modalities included. \textbf{Bold} denotes the best scores for each combination group and dataset. $\ddagger$ denotes the mean of 5 trials.\\
    \footnotesize{SP = Saliency Prediction, FER = Facial Expression Recognition, and GE~=~Gaze~Estimation.}} 
    \resizebox{0.48\textwidth}{!}{%
    \begin{tabular}{c|l|cc|cc}
        \hline
        \multirow{1}{*}{\textbf{Model}} & \multirow{1}{*}{\hphantom{***}\textbf{Cue}} & \multicolumn{2}{c}{MVVA~\citep{liu2020mvva}} & \multicolumn{2}{|c}{FindWho~\citep{xu2018findwho}}\\
        \textbf{Architecture} & SP FER GE & AUCJ$\uparrow$ & NSS$\uparrow$ & AUCJ$\uparrow$ & NSS$\uparrow$ \\
        \hline
        Fusion &\hphantom{*}-\hphantom{**} -\hphantom{**} \checkmark & 0.842 & 0.424 & 0.966 & 1.022 \\
        Integration &\hphantom{*}-\hphantom{**} -\hphantom{**} \checkmark & 0.925 & 0.820 & 0.967 & 1.294 \\
        Fusion &\hphantom{*}-\hphantom{**} \checkmark\hphantom{*} -& 0.938 & 0.846 & \textbf{0.968} & 1.072 \\
        Integration &\hphantom{*}-\hphantom{**} \checkmark\hphantom{*} -& \textbf{0.954} & 1.140 & 0.965 & \textbf{1.691} \\
        Fusion &\hphantom{*}\checkmark\hphantom{*} -\hphantom{**} -& 0.949 & 0.999 & 0.962 & 0.900 \\
        Integration &\hphantom{*}\checkmark\hphantom{*} -\hphantom{**} -  & 0.928 & \textbf{1.171} & \textbf{0.968} & 1.519 \\
        \hline
        Fusion &\hphantom{*}-\hphantom{**} \checkmark\hphantom{*} \checkmark & 0.945 & 0.860 & 0.949 & 0.805 \\
        Integration &\hphantom{*}-\hphantom{**} \checkmark\hphantom{*} \checkmark & \textbf{0.951} & 1.051 & 0.963 & \textbf{1.591} \\
        Fusion &\hphantom{*}\checkmark\hphantom{*} -\hphantom{**} \checkmark & 0.947 & 1.040 & \textbf{0.968} & 1.183 \\
        Integration &\hphantom{*}\checkmark\hphantom{*} -\hphantom{**} \checkmark & 0.932 & 0.580 & 0.951 & 1.489 \\
        Fusion &\hphantom{*}\checkmark\hphantom{*} \checkmark\hphantom{*} -& 0.947 & 0.926 & 0.801 & 0.121 \\
        Integration &\hphantom{*}\checkmark\hphantom{*} \checkmark\hphantom{*} -& 0.943 & \textbf{1.104} & 0.950 & 1.050 \\
        \hline
        Fusion &\hphantom{*}\checkmark\hphantom{*} \checkmark\hphantom{*} \checkmark & 0.952$^\ddagger$ & \textbf{1.352}$^\ddagger$ & 0.969$^\ddagger$ & 1.252$^\ddagger$ \\
        Integration &\hphantom{*}\checkmark\hphantom{*} \checkmark\hphantom{*} \checkmark & \textbf{0.960}$^\ddagger$ & 1.283$^\ddagger$ & \textbf{0.976}$^\ddagger$ & \textbf{2.035}$^\ddagger$ \\
        \hline
    \end{tabular} 
    }
    \label{tab:ablation_study}
\end{table}
To measure the contribution of each social cue, we ablated each modality of the unified integration and fusion models independently and in combination with other modalities. We then trained the models on the MVVA and FindWho datasets separately. For this set of experiments, we report the best of 5 trial scores in~\autoref{tab:ablation_study}, since the variance across trials was large and the mean value was not representative of any of the trained model scores.

When we compared the model's performance with single cue modalities, we observed that the model had the best performance with the Facial Expression Recognition (FER) modality only and the worst with the Gaze Estimation (GE) modality only. When we trained the model with two social cue modalities, we found that on ablating the GE modality, the model performance was negatively impacted with the smaller FindWho dataset but not the larger MVVA dataset. The degradation in the MVVA dataset performance was due to the discrepancy between the number of faces visible in a video and those detected by the GE modality's face detector~\citep{zhang2017ssfd}. The FindWho dataset contains 30/65 (46\%) videos and the MVVA dataset 142/300 (47\%) videos with two or fewer faces. However, when measuring the majority (80\%) of face counts per video frames detected by the GE model, the FindWho dataset had 23/65 (35\%) videos, whereas the MVVA dataset had 211/300 (70\%) videos with two or fewer faces. Moreover, on ablating the saliency prediction or FER modality, there was no impact on the model's performance. We also observed that the inclusion of all social cues consistently yielded the highest performance for both datasets.

\subsection{Multi-Step-Ahead Fixation Prediction}
\label{sec:results_multistepahead}
\begin{figure*}[!t]
\centering
\subcaptionbox{NSS}{\includegraphics[width=0.47\textwidth, trim = 0 1.5em 14em 12.3em, clip=true]{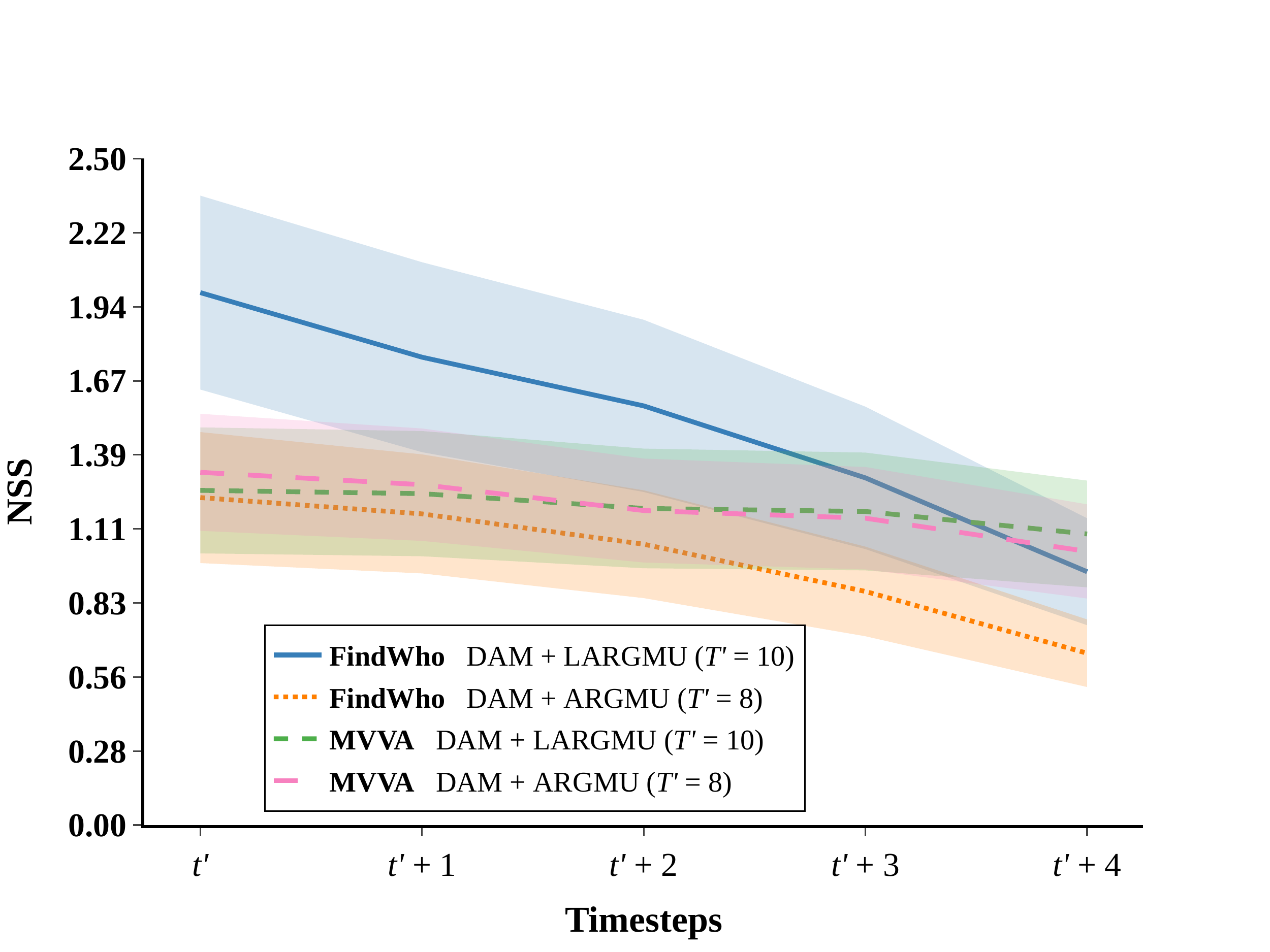}}%
\hfill
\subcaptionbox{AUCJ}{\includegraphics[width=0.47\textwidth, trim= 0 1.5em 14em 12.3em, clip=true]{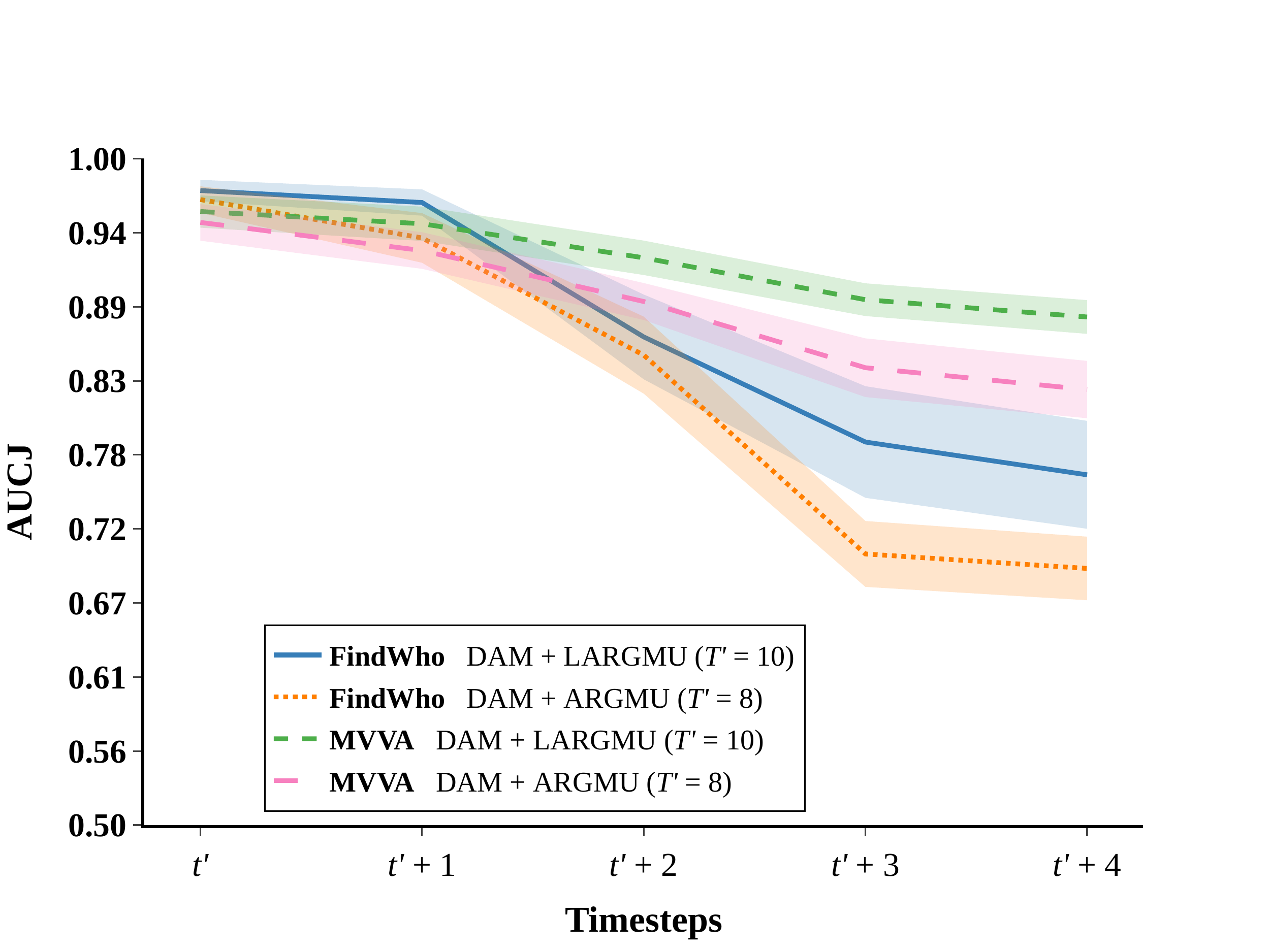}}%
\caption{The two GASP~\citep{abawi2021gasp} variants of the unified model, extended with fixation history modules showing a negative trend in terms of the \textbf{(a)} NSS and \textbf{(b)} AUCJ metrics, as we increase the number of steps ahead. The context size $T'$ for each model is shown in parentheses.}
\label{fig:multistepahead_unified_mvva_findwho}
\end{figure*}

We evaluated the two unified model architecture variants (DAM + LARGMU: integration vs DAM + ARGMU: fusion) as we autoregressively fed the prediction into the fixation history at each timestep. We measured the performance degradation in terms of NSS and AUCJ, over the step-ahead increments $t'$ + $n$~, where $n \in\{1,~\dots~, 4\}$. The evaluation was carried out on the MVVA and FindWho datasets. For each dataset and model architecture, we selected the top-performing model in one out of five training trials. On the MVVA dataset, the fusion model architecture showed a degradation of 22.42\% in NSS and 13.17\% in AUCJ by  $t'$ + 4. The integration model architecture exhibited smaller declines of 13.02\% in NSS and 8.24\% in AUCJ. More significant drops in performance were observed on the FindWho dataset. The fusion model architecture had a 47.53\% decrease in NSS and a 28.55\% decrease in AUCJ. The integration model architecture exhibited the highest declines with 52.42\% in NSS and 21.85\% in AUCJ by  $t'$ + 4.  The results indicate a trend of performance reduction over extended prediction horizons across all model variants due to the accumulation of errors. Models trained and evaluated on the FindWho dataset, particularly, showed more significant degradation.~\autoref{fig:multistepahead_unified_mvva_findwho} illustrates that models trained on the small FindWho dataset are less robust in predicting multiple steps ahead than when trained on the larger MVVA dataset, as indicated by the steeper negative trends in AUCJ and NSS. In~\autoref{tab:multistepahead_unified_mvva_findwho}, we also observe that the late integration architecture tends to outperform early fusion over longer horizons.
\begin{table*}[!ht]
    \centering
    \caption{Multi-step-ahead predictions using unified models fine-tuned on the small FindWho~\citep{xu2018findwho} and large MVVA~\citep{liu2020mvva} datasets. All models are based on variants of GASP~\citep{abawi2021gasp} with the additional fixation history modules. The context size $T'$ for each model is shown in parentheses. \textbf{Bold} denotes the best scores for each step.}
    \resizebox{\textwidth}{!}{%
    \begin{tabular}{l|l|cc|cc|cc|cc|cc}
        \hline
        \multirow{2}{*}{\textbf{Model Architecture}} & \multirow{2}{*}{\textbf{Dataset}} & \multicolumn{2}{c|}{$\boldsymbol{t'}$} & \multicolumn{2}{c|}{$\boldsymbol{t' + 1}$} & \multicolumn{2}{c|}{$\boldsymbol{t' + 2}$} & \multicolumn{2}{c|}{$\boldsymbol{t' + 3}$} & \multicolumn{2}{c}{$\boldsymbol{t' + 4}$} \\
        && AUCJ$\uparrow$ & NSS$\uparrow$ & AUCJ$\uparrow$ & NSS$\uparrow$ & AUCJ$\uparrow$ & NSS$\uparrow$ & AUCJ$\uparrow$ & NSS$\uparrow$ & AUCJ$\uparrow$ & NSS$\uparrow$ \\
        \hline
        Fusion: DAM + ARGMU $(T'=8)$& FindWho& 0.969 & 1.252 & 0.941 & 1.190 & 0.853 & 1.074 & 0.703 & 0.893 & 0.693 & 0.657 \\
        Integration: DAM + LARGMU $(T'=10)$& FindWho& \textbf{0.976} & \textbf{2.035} & \textbf{0.967} & \textbf{1.788} & 0.866 & \textbf{1.603} & 0.787 & \textbf{1.327} & 0.763 & 0.969 \\
        \hline
        Fusion: DAM + ARGMU $(T'=8)$& MVVA& 0.952 & 1.352 & 0.931 & 1.305 & 0.893 & 1.206 & 0.843 & 1.176 & 0.827 & 1.049 \\
        Integration: DAM + LARGMU $(T'=10)$& MVVA& 0.960 & 1.283 & 0.951 & 1.271 & \textbf{0.926} & 1.214 & \textbf{0.894} & 1.202 & \textbf{0.881} & \textbf{1.116} \\
        \hline
    \end{tabular}
    }
    \label{tab:multistepahead_unified_mvva_findwho}
\end{table*}

\section{Discussion}
\label{sec:discussion}

In \autoref{sec:results_architecture_dataset_1v1_1vinf}, we analyzed the individual models with two architectures (late integration vs early fusion) on the FindWho and MVVA datasets. We observed that when training and evaluating on the FindWho dataset, the fusion architecture performed on par with the integration architecture in terms of the NSS and AUCJ. Additionally, the fusion architecture exhibited lower variance across videos per participant when measuring scanpath correspondence between each individual model's prediction and its own (\textit{1~vs~1}) as well as other observers' (\textit{1~vs~infinity}) ground-truth data. This suggests that the fusion architecture is more stable than the integration variant. Another important finding was that all model architectures performed significantly better on \textit{1~vs~1} than on \textit{1~vs~infinity}. This indicates that each model learned the scanpath of one individual, and did not simply represent the group's attention. On the contrary, saliency prediction models would score significantly better on \textit{1~vs~infinity} (in this case, \textit{infinity~vs~infinity}) than on \textit{1~vs~1} (in this case, \textit{infinity~vs~1}) evaluations~\citep{bylinskii2019eval}.

Moreover, the integration architecture's stability was contingent on the dataset. When a model architecture performs well on the FindWho dataset---which has fewer videos than the MVVA dataset but a similar distribution---it suggests that the model effectively infers patterns of attention that are shared or common across the majority of observers. This is related to understanding universal attention, where most observers converge in their attention patterns due to the limited video variability. On the other hand, with the MVVA dataset having a larger set of videos, there's a higher likelihood for individual variations in attention patterns to emerge. Thus, a model's success on MVVA can indicate its ability to discern and adapt to these individualized attention behaviors. This aligns with the notion of personalized attention, where the attention patterns might be more specific to individual inclinations, experiences, or abilities. Training and evaluating on the MVVA dataset resulted in significantly lower variance across videos compared to FindWho, in terms of NSS and AUCJ on both \textit{1~vs~1} and \textit{1~vs~infinity} evaluations. As the number of observation videos increases, scanpath patterns begin to align among individual viewers. This suggests that although the models are exposed to more samples of personalized attentional behavior, on longer exposure to stimuli, universal attention exerts greater influence on viewing patterns. Consequently, integration models trained on the MVVA dataset underperform counterparts trained on the FindWho dataset, however, the variance across videos is also significantly lower. The lower variance in the case of the MVVA dataset is both due to ``regression toward the mean'' and bottom-up saliency affecting participants in relatively equal proportions, given its larger number of videos.

Training a separate model for each individual is costly in terms of computational resources, making it an impractical approach for training models on larger datasets with many observers. To overcome such a limitation, we devised a unified training approach, whereby the model is exposed to scanpaths of all individuals during training. Each scanpath is trained in the exact manner as the individual model, with the main difference being the sampling strategy. The individual models are only fed samples from a single observer during training. However, the unified model samples a random observer's scanpath, and is trained with the corresponding stimuli and fixation histories.

In ~\autoref{sec:results_interaction_effects}, we compared unified and individual models. The unified model performed significantly better than the individual models. This improvement can be attributed to the fact that the unified model is trained on data from all observers, subjecting it to greater variability in the samples. As a result, the unified model is exposed to a larger spectrum of traits relating to universal attention. More importantly, the unified model, regardless of the architecture, can predict different scanpaths given the same stimuli, conditioned only on the fixation history. This is evident from the scores being comparable to the individual models, which represent the baseline for acceptable performance. According to these results, we infer that the fixation history is a sufficient prior since:
\begin{enumerate}
    \item Scores of the unified model are on par or better than those of the individual models.
    \item The fixation history is the only prior available to the model for it to differentiate scanpath trajectories. Without it, the model would generate arbitrary scanpaths, consequently performing significantly worse than the individual models.
\end{enumerate}

In~\autoref{sec:results_socialcueablation}, our results showed that including all social cue modalities improved the performance of our models. Moreover, ablating the gaze estimation modality degraded the performance of both model architectures when trained and evaluated on the smaller FindWho dataset, yet had negligible effect on the larger MVVA dataset. The gaze estimations were represented as gaze cones, superimposed on the face positions of actors visible in a video frame. These cones were oriented according to the estimated gaze direction of each actor, after which they were normalized and aggregated. Having more than two gaze cones---more than two faces detected---in any frame distorts the gaze estimation representation. We therefore assume that occurrences of two or fewer face detections are optimal for gaze estimation. The MVVA dataset was found to result in more detections (70\%) of videos with two or fewer faces than the FindWho dataset (35\%), even though both datasets contained \around 46\% videos with two or fewer faces. This implies that gaze estimation representations of the MVVA samples were inaccurate, resulting in our models relying less on that social cue.

Predicting one step ahead for each scanpath is useful only when the model has access to the ground-truth fixation history of an individual. In practice, however, this requirement renders a model unusable for most applications. K\"ummerer~and~Bethge~\cite{kuemmerer2021scanpath} present an evaluation framework that addresses this limitation. By feeding the output of a model recursively into its fixation history module, we can evaluate the model for multiple steps ahead without relying on the ground-truth fixations (as input) beyond the initial steps. This form of evaluation informs us on the performance of a model during inference, and how likely the output predictions are to diverge from the ground-truth over multiple steps.

In~\autoref{sec:results_multistepahead}, we followed a similar approach in evaluating our unified models. Results indicated that although the unified integration model was not the most performant under one-step-ahead evaluation, it exhibited higher robustness compared to the fusion architecture as the number of steps ahead increased. This result implies that the late integration of cue representations, namely, applying recurrence before gating, is more beneficial for learning sequences, and vice versa. Moreover, training on the MVVA dataset stabilized the predictions of a model over longer horizons, as shown by the least reduction in scores over multiple steps ahead. We hypothesize that MVVA-trained models are exposed to more samples and therefore are less affected by the accumulation of errors as the number of steps ahead increases due to higher variability in the samples.

Our findings, however, do not imply that our unified models could identify individuals from their scanpath trajectories. Individuals' scanpaths can be separated into clusters~\citep{coutrot2018scanpath}, indicating that although they are distinct from the group, they do not differ significantly with respect to all other individuals independently. Since the fixation history---covering no longer than a few seconds of low-dimensional data---allows the unified model to perform on par with the individual models, we deduce that the viewing patterns of individuals are not significantly different. Differences in past scanpaths do influence subsequent fixations, but the unified model still ensures privacy because it does not tie predictions to specific observers, unlike the individual models where the observer is known a priori.

Overall, we introduced a mechanism for integrating fixation history into dynamic video models, designed to personalize scanpaths to a specific observer. A comparative analysis with existing models could be considered for measuring performance, yet many factors limit our ability to do so. For instance, an essential component for enabling the personalization of scanpaths is the fixation history. However, most existing dynamic models do not include a fixation history module. Moreover, dynamic models that encode fixation history~\citep{naas2020functional,xu2018gaze} are commonly designed for scanpath prediction in 360$^\circ$ videos, and adopt approaches such as image patching---splitting the visual frames into smaller segments to simulate foveation---that make them unsuitable for the dataset videos used in this work, due to their limited resolution. Additionally, adapting these models would require major modifications to their architectures and tasks, ranging from introducing a fixation history module, performing hyperparameter optimization, to retraining or fine-tuning on the FindWho and MVVA datasets. Consequently, any adapted models would deviate from their implementations, resulting in the creation of new models. Given these constraints, we evaluated our approach using our own models with different integration architectures and ablation studies, providing baselines for future comparisons.

\section{Conclusion}
\label{sec:conclusion}
Studies on goal-directed human attention indicate insignificant differences in scanpaths among individuals~\citep{yang2020irl,yang2022targetabsent}. Contrary to such findings, free-viewing entails complex top-down influences, resulting in significant variance across viewing patterns among different observers~\citep{bylinskii2019eval, chen2021scanpathvqa, coutrot2018scanpath}. We introduced a framework for predicting and evaluating individual scanpaths in social videos. We focused mainly on social videos due to their complexity, offering varied audiovisual cues and interactions for our scanpath prediction framework to analyze. Our main finding was that the introduction of fixation history into the model was a sufficient prior for allowing a single unified model to predict scanpaths. However, this does not imply that the unified models can identify individual observers through their scanpath trajectories.

\textcolor{black}{Our model is limited to predicting scanpaths in social videos. However, even in such settings, human faces---from which we detect and extract social cues---are at times occluded or not visible throughout the video. Moreover, our model's predictions are deterministic. When the two input samples consisting of the same fixation history and spatiotemporal representations are fed into our model, it predicts identical priority maps. However, multiple viewings of a scene by the same observer do not necessarily result in the same scanpath trajectories.}

In future work, we will extend our dynamic scanpath prediction model to handle the non-deterministic nature of gaze. We intend to integrate techniques from static scanpath prediction models, particularly Generative Adversarial Imitation Learning~\citep{yang2020irl} and Reinforcement Learning~\citep{chen2021scanpathvqa}, adapting these methods for dynamic video inputs. We will also study the effect of introducing further social cue representations into the model, such as full body gestures, biological motion, intonation, and prosodic features. \textcolor{black}{Another aspect to be examined is the effect of varying the starting timestep for the context window. In our scanpath prediction model, although a prediction is made for the next fixation, not every last frame preceding the prediction is equally challenging. For instance, if the context contains relatively few gaze shifts, the prediction task is inherently easier than when the context includes multiple shifts. This variability, determined by where the context begins and ends, will be analyzed to better understand its impact on performance.}

\section*{CRediT authorship contribution statement}
\textbf{Fares Abawi:} Conceptualization, Methodology, Software, Validation, Formal analysis, Investigation, Resources, Data curation, Writing -- original draft, Writing -- review \& editing, Visualization. 
\textbf{Di Fu:} Conceptualization, Formal analysis, Data curation, Writing -- original draft, Writing -- review \& editing, Visualization. 
\textbf{Stefan Wermter:} Resources, Writing -- review \& editing, Supervision, Funding acquisition.

\section*{Declaration of competing interest}
The authors declare that they have no known competing financial interests or personal relationships that could have appeared to influence the work reported in this paper.

\section*{Data availability}
Data will be made available on request. 

Code and models available at: 

\url{https://github.com/knowledgetechnologyuhh/gasp}

\section*{Acknowledgements}
The authors would like to thank Dr. Cornelius Weber (University of Hamburg) for his valuable feedback and insights on teacher forcing and sequential learning. The authors gratefully acknowledge partial support from the German Research Foundation DFG under project CML~(TRR~169). 
 \bibliographystyle{elsarticle-num} 
 \bibliography{submission}

\end{document}